\documentclass[sigplan,nonacm]{acmart}
\makeatletter
\def\@ACM@checkaffil{% Only warnings
    \if@ACM@instpresent\else
    \ClassWarningNoLine{\@classname}{No institution present for an affiliation}%
    \fi
    \if@ACM@citypresent\else
    \ClassWarningNoLine{\@classname}{No city present for an affiliation}%
    \fi
    \if@ACM@countrypresent\else
        \ClassWarningNoLine{\@classname}{No country present for an affiliation}%
    \fi
}
\makeatother

\usepackage[]{hyperref}
\usepackage{amsmath}
\usepackage{subcaption}
\usepackage{graphicx}
\usepackage{xspace}
\usepackage{ifthen}
\usepackage{xcolor}
\usepackage{array}
\usepackage{cleveref}
\usepackage{svg}
\usepackage{tikz}

\usepackage{xparse}
\usepackage{multirow}
\usepackage{booktabs}
\usepackage{arydshln}
\usepackage{enumitem}
\usepackage{algorithm}
\usepackage{algpseudocode}

\newcommand{\sysname}{\textsc{Niyama}\xspace}
\newcommand{\sref}[1]{(\S\ref{#1})}

\newcommand{\todo}[1]{\textcolor{red}{#1}}
\newcommand{\jheading}[1]{\vspace{0.05in}\noindent\textbf{#1.}}
\newcommand{\code}{Azure-Code\xspace}

\newcommand{\fcfs}{Sarathi-FCFS\xspace}
\newcommand{\edf}{Sarathi-EDF\xspace}
\newcommand{\srpf}{Sarathi-SRPF\xspace}
\definecolor{darkgreen}{rgb}{0.0, 0.5, 0.0}

\newcommand*\circled[1]{\tikz[baseline=(char.base)]{
\node[shape=circle,draw,inner sep=0.8pt, fill=gray!20] (char) {\small #1};}}

\algrenewcommand\algorithmiccomment[1]{\hfill// \textnormal{#1}}

\makeatletter
\renewcommand{\sectionautorefname}{\S\@gobble}
\renewcommand{\subsectionautorefname}{\S\@gobble}
\renewcommand{\subsubsectionautorefname}{\S\@gobble}
\makeatother

\usepackage{ifthen}
\newboolean{publicversion}
\setboolean{publicversion}{false}

\ifthenelse{\boolean{publicversion}}{
    \newcommand{\todo}[1]{}

    \newcommand{\grumbler}[3]{}
    \newcommand{\jm}[1]{}
    \newcommand{\nk}[1]{}
    \newcommand{\rr}[1]{}
    \newcommand{\kg}[1]{}
    
}
{
    \newcommand{\grumbler}[3]{\xspace\textcolor{#3}{\bf #1: #2}}
    \newcommand{\jm}[1]{\grumbler{Jayashree}{#1}{magenta}}
    \newcommand{\nk}[1]{\grumbler{Nipun}{#1}{red}}
    \newcommand{\rr}[1]{\grumbler{Ram}{#1}{teal}}
    \newcommand{\kg}[1]{\grumbler{Kanishk}{#1}{violet}}
    
}

\usepackage[absolute]{textpos}

\begin{document}

\title{\sysname: Breaking the Silos of LLM Inference Serving}

\author{
    Kanishk Goel, \hspace{3pt}
    Jayashree Mohan, \hspace{3pt}
    Nipun Kwatra, \hspace{3pt}
    Ravi Shreyas Anupindi, \hspace{3pt}
    Ramachandran Ramjee 
    \vspace{6pt}\\
    }
    \affiliation{
    \institution{Microsoft Research India}
    %\city{Bangalore}
    %\country{India}
    \vspace{3em}
    }

\pagestyle{plain}

\begin{abstract}
The widespread adoption of Large Language Models (LLMs) has enabled diverse applications with very different latency requirements. Existing LLM serving frameworks rely on siloed infrastructure with coarse-grained workload segregation --- interactive and batch --- leading to inefficient resource utilization and limited support for fine-grained Quality-of-Service (QoS) differentiation. This results in operational inefficiencies, over-provisioning and poor load management during traffic surges.

We present \sysname, a novel QoS-driven inference serving system that enables efficient co-scheduling of diverse workloads on shared infrastructure. \sysname introduces fine-grained QoS classification allowing applications to specify precise latency requirements, and dynamically adapts scheduling decisions based on real-time system state. Leveraging the predictable execution characteristics of LLM inference, \sysname implements a dynamic chunking mechanism to improve overall throughput while maintaining strict QoS guarantees. Additionally, \sysname employs a hybrid prioritization policy that balances fairness and efficiency, and employs selective request relegation that enables graceful service degradation during overload conditions.
Our evaluation demonstrates that \sysname increases serving capacity by 32\% compared to current siloed deployments,while maintaining QoS guarantees. Notably, under extreme load, our system reduces SLO violations by an order of magnitude compared to current strategies.

\end{abstract}

\maketitle

\section{Introduction}

%The explosive growth of large language models (LLMs) has transformed the landscape of artificial intelligence applications, enabling

% Intro - need for QoS
Large language models (LLMs) have transformed applications across diverse domains including conversational assistants, coding assistants, content generation, and summarization. These applications can have very different latency requirements --- for example, autocomplete coding assistants demand responses within milliseconds, while summarization tasks can reasonably tolerate longer latencies.  As LLM deployments scale to serve billions of users and diverse applications, inference serving systems must efficiently handle this diverse spectrum of latency requirements while ensuring high GPU utilization.

%Current deployments of LLM inference systems face significant limitations in resource utilization, quality of service (QoS) guarantees, and graceful degradation under load. These challenges grow increasingly acute as LLMs become foundational infrastructure for numerous applications with varying requirements.

\begin{figure}[t]
    \centering
    \begin{subfigure}[b]{0.5\linewidth}
        \includegraphics[width=\linewidth]{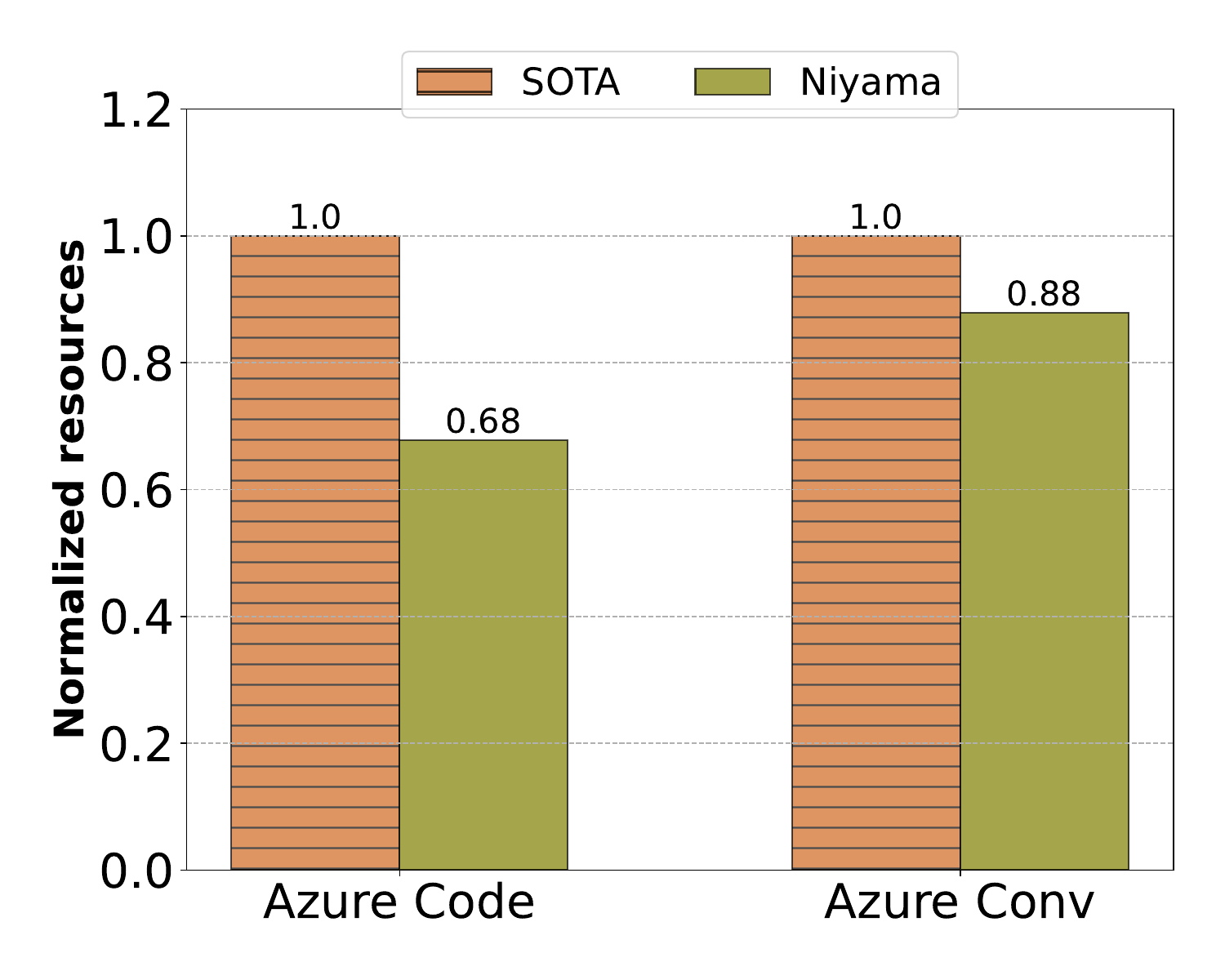}
    \end{subfigure}
\hfill
    \begin{subfigure}[b]{0.4\linewidth}
        \includegraphics[width=\linewidth]{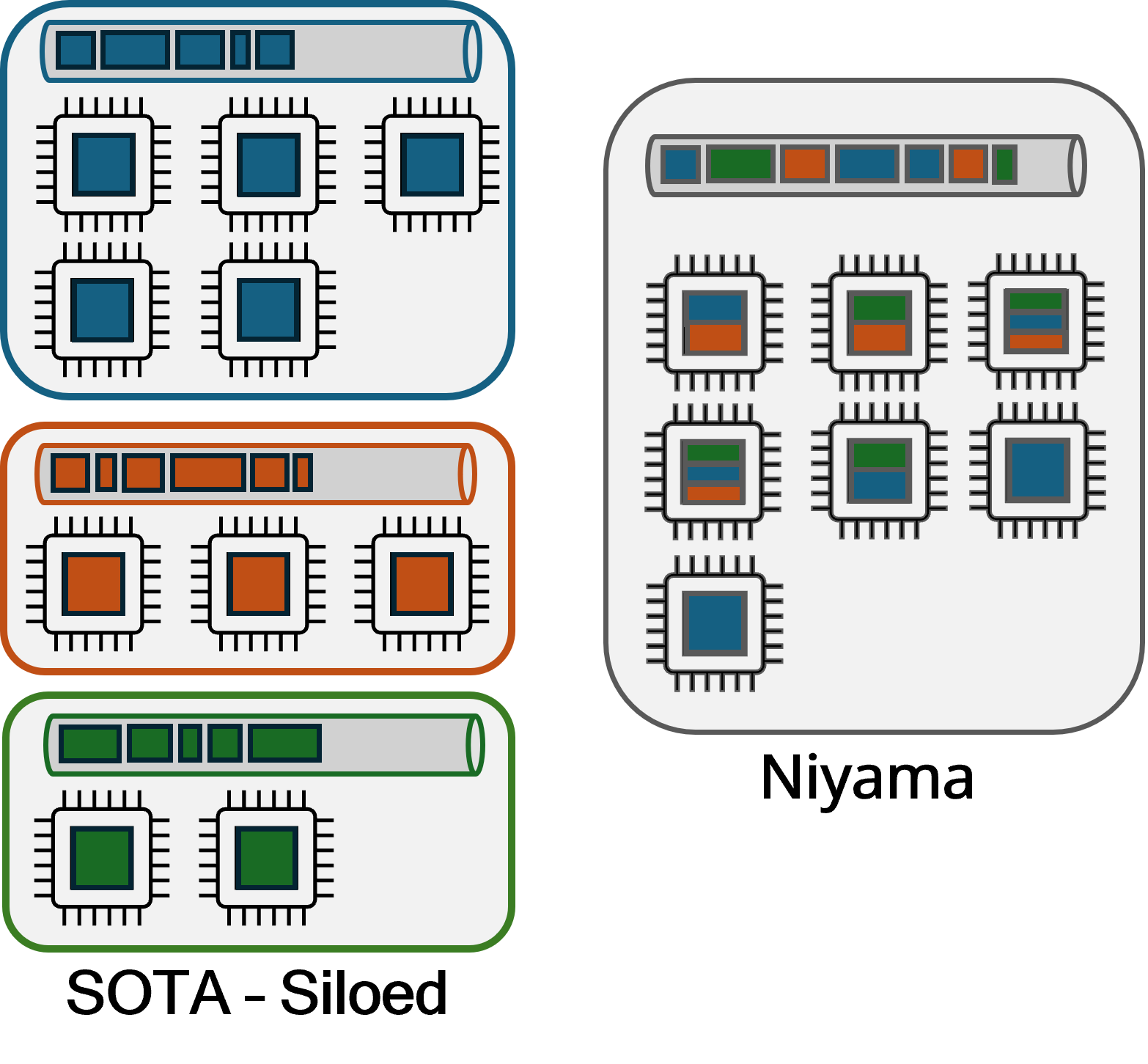}
    \end{subfigure}

    \begin{subfigure}[b]{0.48\linewidth}
        \includegraphics[width=\linewidth]{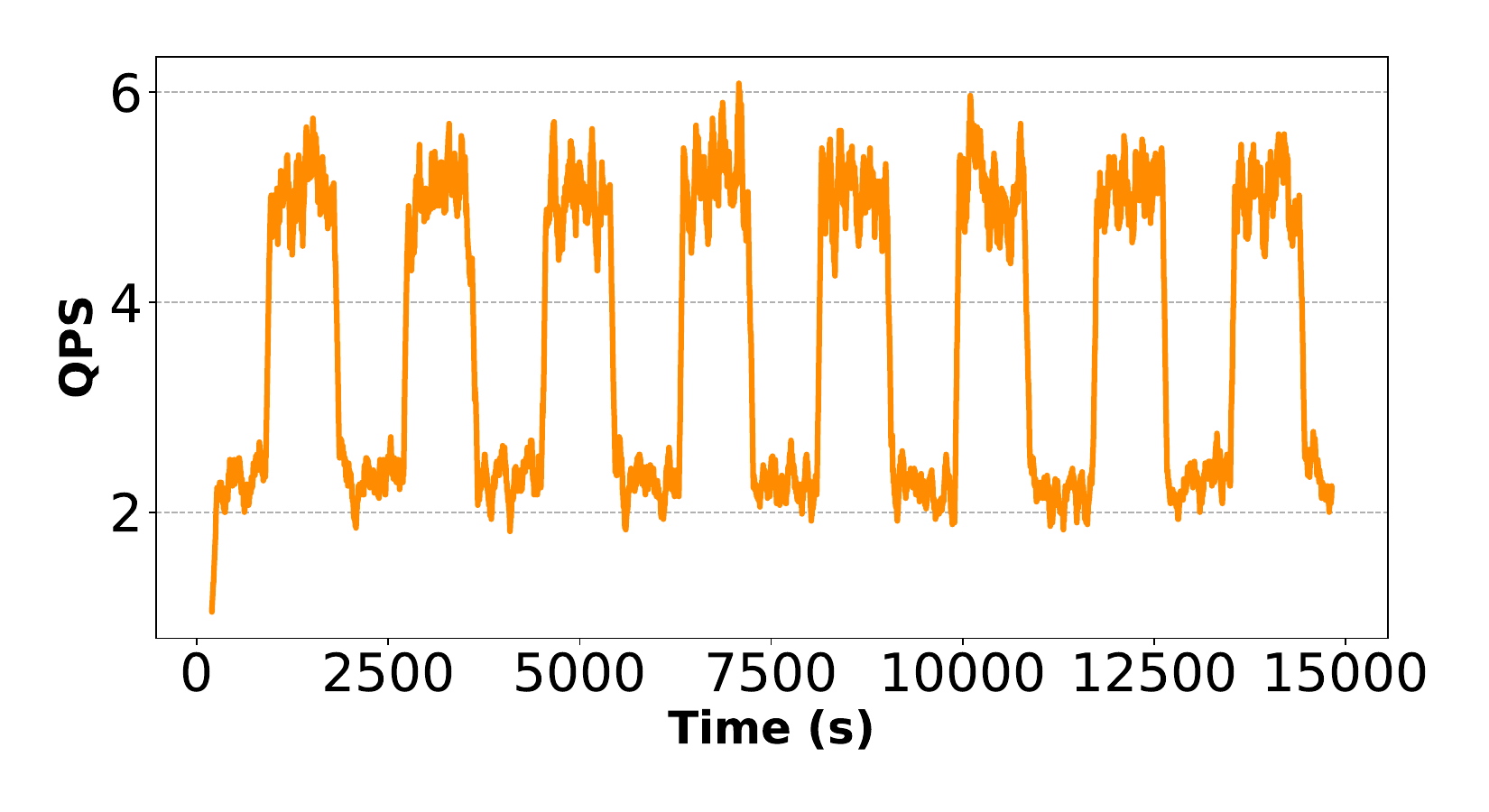}
    \end{subfigure}
\hfill
    \begin{subfigure}[b]{0.48\linewidth}
        \includegraphics[width=\linewidth]{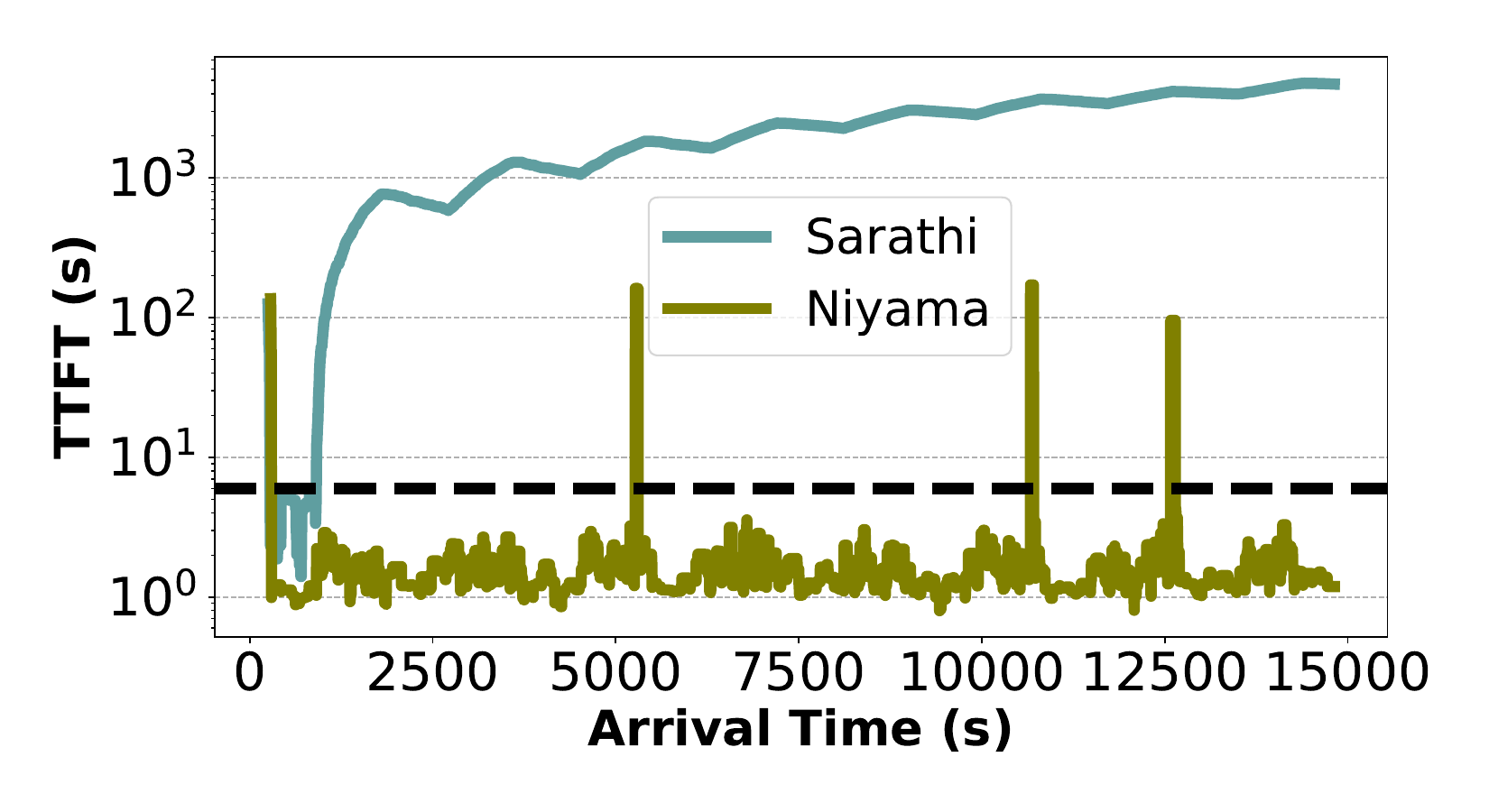}
    \end{subfigure}
    
    \caption{Efficiency of \sysname under uniform load and transient overload conditions. (top left) Normalized GPUs needed to serve a fixed traffic load while meeting the QoS targets of requests divided equally among 3 QoS tiers. Across two datasets, \sysname improves efficiency by 12-32\% compared to the State-of-the-art (SOTA) Sarathi-Serve~\cite{sarathi2024} siloed deployments.
    (top right) Illustration of \sysname co-scheduling vs current siloed deployments. (bottom left) Bursty overload scenario. (bottom right) \sysname maintains low latency while SOTA scheduling succumbs to cascading deadline violations under such bursty loads.}
    \label{fig:intro:banner}
    \vspace{-1em}
\end{figure}

% Talk about the deployment scenario today
Current LLM serving solutions primarily adopt a coarse-grained categorization, segregating requests into two broad service classes: latency-sensitive interactive applications, and throughput-oriented batch processing, and serving them independently~\cite{jaiswal2025serving}.
%\footnote{Based on deployment information from a large cloud provider X}.
Interactive requests are typically served with smaller prefill chunks~\cite{sarathi2024} to minimize latency, but that can result in relatively higher operational costs due to reduced throughput (e.g., 28\% lower as shown in Figure~\ref{fig:chunk-perf}). Batch requests, on the other hand, employ larger chunks to achieve higher throughput  as latency is not a constraint. This siloed deployment, however, creates other inefficiencies: it leads to significant GPU resource under-utilization, as workload demands fluctuate across the two classes. 
%Consequently, providers resort to costly over-provisioning in each silo to guarantee service levels. 
Moreover, such partitioning inhibits the introduction of more QoS classes with fine-grained latency requirements, as doing so further exacerbates the partitioning inefficiencies.

Furthermore, current inference systems struggle under load fluctuations and overload conditions. Typical scheduling mechanisms such as first-come-first-served (FCFS) indiscriminately delay all incoming requests under overload, degrading user experience across the board. Alternatively, simplistic throttling approaches reject all new incoming requests when reaching capacity, ignoring their QoS requirements or relative priorities. Neither strategy adequately manages the complex trade-offs between throughput, latency, and fairness during such demand surges.

%Current inference systems also handle load surges poorly, defaulting to one of two unsatisfactory approaches when faced with traffic exceeding capacity. Many systems employ first-come-first-served (FCFS) scheduling that indiscriminately increases queuing delays for all requests, and effectively denies service to all users. Alternatively, some deployments implement simplistic throttling mechanisms that reject incoming requests when capacity is reached, without considering their relative importance or the specific QoS requirements of different applications. Both these approaches fail to properly manage the trade-offs involved in providing consistent service under varying load conditions.

% Introduce Niyama
In this paper, we present \sysname, a QoS-driven LLM inference serving system that addresses these limitations through two key ideas. First, \sysname supports fine-grained QoS classes which allows applications to precisely specify their latency requirements. Multiple QoS classes are served efficiently by {\it co-scheduling requests with diverse QoS targets on a shared rather than siloed infrastructure}. Second, \sysname implements a hybrid prioritization and an eager relegation policy that {\it allows graceful service degradation during overload conditions}. ~\Cref{fig:intro:banner} compares \sysname to state-of-the-art Sarathi-Serve~\cite{sarathi2024} siloed deployment, demonstrating significant performance improvements.
%\todo{Probably chunking is introduced abruptly here. Must do this better}

Efficiently supporting multiple QoS classes on a shared serving instance poses significant challenges. One approach is to use the smallest chunk size necessary to meet the latency constraint of the strictest QoS class on all serving instances. However, this would result in low throughput~\cite{sarathi2024} and high cost for all service classes. Instead, \sysname leverages the unique execution characteristics of LLM inference -- particularly the distinct prefill and decode phases and the inherent predictability of the prefill phase -- to dynamically adjust chunk sizes based on the observed system state and individual QoS targets. Co-serving multiple QoS classes allows us to exploit {\it deadline slack} of requests with relaxed latency requirements to schedule bursts of larger chunk sizes, thereby increasing throughput opportunistically. Our evaluations demonstrate that \sysname achieves up to 32\% higher effective serving capacity compared to current siloed deployments, while consistently meeting QoS guarantees across all request categories. 

For managing overload conditions gracefully, \sysname employs a hybrid prioritization and an eager relegation policy. Simple overload handling approaches like shortest-job-first (SJF) manage overload by prioritizing short requests. This helps reduce load due to the quadratic dependence of request length on LLM system load~\cite{attentionpaper}. However, SJF neglects the QoS requirements of longer jobs, leading to SLO violations even at low load (Figure~\ref{fig:motivation_sched}). On the other hand, Earlier Deadline First (EDF) scheduling is optimal under low load but suffers excessive violations even when load is slightly higher than capacity. Thus, \sysname introduces a hybrid policy that smoothly interpolates between EDF and SJF, allowing deployments to minimize SLO violations across both low and high load. Additionally, \sysname proactively employs \textit{eager relegation}, selectively degrading service for a small subset of requests to ensure stable performance, even under extreme load conditions. In multi-QoS scenarios, \sysname leverages application-provided hints about request importance, such as whether a request originates from a free or paid tier, to perform relegation. This ensures that lower-priority requests are affected first during overload conditions, allowing the system to maintain QoS for the majority of high-priority requests.
%\todo{can we add some line to the effect that in multi-qos scenario, we can leverage priorities to do this relegation?}
Our evaluations show that during significant overload scenarios (50\% above capacity), \sysname consistently meets latency targets for over 95\% of requests, translating into substantial cost savings and enhanced user experience across diverse applications relying on LLM infrastructure.

%During overload, instead of throttling all requests, one simple strategy is to do shortest-job-first (SJF) scheduling. SJF effectively delays all long requests. Since there is a quadratic dependence of request length on LLM system load~\cite{attentionpaper}, SJF manages overload better than FCFS. However, SJF is unfair to long requests, implicitly assuming that only short requests are important. Instead, \sysname  uses a novel hybrid policy that creates a continuum between FCFS and SJF scheduling approaches, allowing deployments to navigate the inherent trade-off between fairness and efficiency. Additionally, \sysname makes {\it eager load-shedding} decisions and degrades service gracefully by selectively delaying a small set of requests to maintain consistent performance for a high percentile of requests in the system even under extreme load. During significant overload conditions (XX\% load over capacity), \sysname ensures that over 95\% of  requests meet their latency targets by selectively degrading service for the rest. These improvements translate into substantial cost savings and more consistent user experience across the diverse applications that increasingly rely on LLM infrastructure.

Our work makes the following key contributions:

\begin{enumerate}
    %\item We introduce fine-grained QoS classes that allows applications to express their precise latency requirements, enabling more efficient resource sharing across diverse workloads.
    \item We develop a QoS-aware adaptive scheduling algorithm that exploits the unique characteristics of LLM inference to co-schedule requests belonging to multiple QoS classes on shared infrastructure, improving throughput while maintaining latency guarantees.
    \item We design and implement a hybrid prioritization and eager relegation policy that minimizes SLO violations under both optimal load and overload conditions.
    \item We evaluate \sysname on various workloads and scenarios, demonstrating up to 32\% higher serving capacity while meeting QoS guarantees compared to current approaches.
\end{enumerate}

The rest of the paper is structured as follows. \sref{sec:motivation} outlines the need for QoS-based serving systems and \sref{sec:design-impl} details the architecture and implementation of \sysname. \sref{sec:evaluation} presents our evaluation methodology and results.

\section{Background and Motivation}
\label{sec:motivation}

% inference and sarathi scheduling background
% background on how LLM inference works. Today scheduling is typically done using chunked prefills - cite sarathi. Briefly explain it
\subsection{LLM Inference}
Large language model (LLM) inference is fundamentally different from traditional computing workloads, characterized by two distinct computational phases that significantly impact system design: the prefill and decode stages. During the prefill phase, the entire input prompt is processed simultaneously, making it  computationally intensive. The subsequent decode phase generates output tokens auto-regressively, with each token's generation depending on the previously generated tokens.

\jheading{Scheduling} In this work, we assume co-located LLM inference scheduling as seen in popular serving frameworks like vLLM~\cite{vllmsosp} and SGLang~\cite{zheng2024sglangefficientexecutionstructured} where prefills and decodes of a request are executed on the same replica using chunked prefills~\cite{sarathi2024} for better serving efficiency. Chunked prefills split a prefill request into equal-sized chunks, allowing for efficient batching and scheduling without pausing ongoing decodes. This approach helps balance the trade-off between throughput and latency, and is used as a standard scheduling practice in production systems~\cite{anyscale}.

% Latency metrics
%Talk about two classes of applications - latency sensitive, user facing and non user-facing. In the first category we care about TTFT and TPOT - explain each. In the second, we care about TTLT only.
\jheading{Latency metrics} LLM inference encompasses three primary latency metrics, which serve as critical performance indicators across different application types:

\begin{enumerate}[itemsep=0pt, topsep=0pt, left=0pt]
    \item \jheading{Time to First Token (TTFT)} This metric captures the initial response latency, measuring the duration from request submission to generating the first output token. For interactive applications like chatbots, coding assistants, or real-time conversational interfaces, TTFT is crucial as it directly influences user perception of system responsiveness.
    \item \jheading{Time Between Tokens (TBT)} This metric measures the interval between the generation of consecutive output tokens of a request, and affects the overall perceived fluidity of the response which is particularly important for interactive applications where users expect a smooth, uninterrupted stream of generated content.
    
    \item \jheading{Time to Last Token (TTLT)} This metric focuses on the total time required to complete the entire generation process. TTLT is particularly relevant for non-interactive, batch-oriented applications such as document summarization, comprehensive research analysis, or offline content generation. In these scenarios, the overall completion time matters more than the speed of initial response or token-by-token generation.
\end{enumerate}

\noindent
The application's nature determines which of these metrics take priority. User-facing, interactive applications critically depend on both TTFT and TBT, as these metrics directly impact user experience and perceived system responsiveness. In contrast, non-interactive applications primarily concern themselves with TTLT, prioritizing the total time to generate a complete output over the speed of initial token generation.

\begin{comment}
\begin{enumerate}
    \item Latency-Sensitive, User-Facing Applications. These applications prioritize metrics such as TTFT(Time-To-First-Token) and TBT(Time-Between-Token). For a given request, TTFT measures the latency of generating the first output token from the moment a request arrives in the system. This metric reflects the initial responsiveness of the model. TBT on the other hand measures the interval between the generation of consecutive output tokens of a request, and affects the overall perceived fluidity of the response. When system is under high load,low throughput can lead to large scheduling delays and consequently higher TTFT. Both metrics are crucial for providing a responsive user experience.

    \item Non-User-Facing Applications. For these applications, the primary concern is Time To Last Token (TTLT), which measures the total time taken to generate the complete output. These applications can tolerate higher latencies as they are not directly interacting with users.
\end{enumerate}
\end{comment}

% add a figure here to show silo'd deployments
%Current industrial practice partitions LLM workloads into primarily two siloed clusters - latency-sensitive interactive requests and batch requests each with dedicated GPU fleets.
%mechanisms of overload management - no graceful degradataion, techniques like user rate limiting, prioritizing short requests are unfair and again not application-aware or graceful. Find public refernces for QoS and overload management of LLM inference
\subsection{Production Deployment Landscape}

Due to the fundamental differences in workload characteristics and performance requirements between these application types, current industrial practices for LLM inference deployment predominantly employ %\todo{can we add any citations for this?}
a siloed infrastructure model~\cite{jaiswal2025serving}, maintaining two distinct GPU clusters:
(1) a dedicated fleet for latency-sensitive, interactive requests, and
(2) a separate cluster for batch processing and background jobs.

%This approach emerges from the fundamental differences in workload characteristics and performance requirements between these application types.

\begin{figure*}[t]
    \centering
\includegraphics[width=0.98\linewidth]{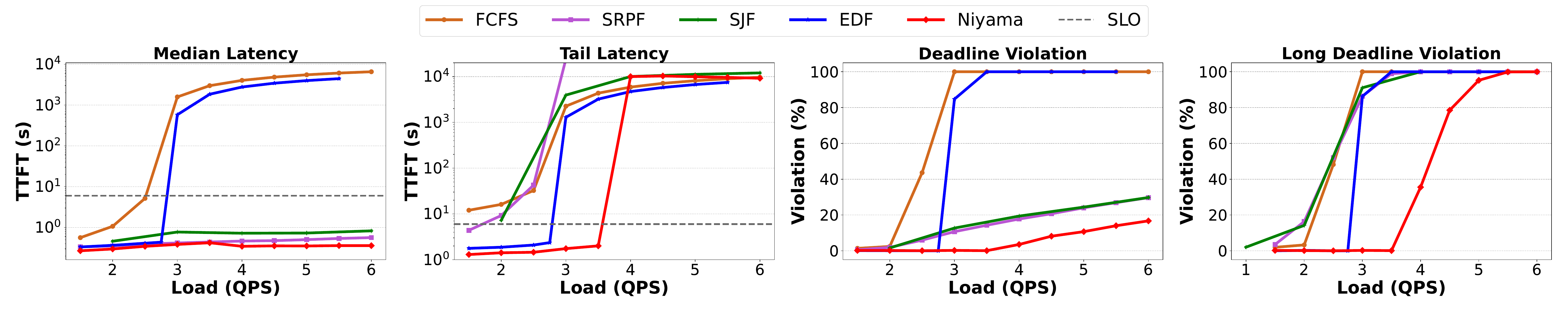}
    \caption{Comparison of traditional  policies for multi-SLA scheduling. The graphs plot the latency and violations in the strictest QoS class. FCFS breaks down very quickly because urgent requests can be blocked by non-urgent ones. Deadline-aware policies like EDF are better than FCFS, but cannot gracefully degrade at high loads because of intense queue buildup. SJF/SRPF on the other hand can maintain QoS in the median case but  violates SLOs of majority of long jobs even at a low load of 2.5 QPS. \sysname{} interpolates smoothly between SJF and EDF and minimizes violations across all load conditions.}
    \label{fig:motivation_sched}
\end{figure*}

\jheading{Overload management} 
When faced with traffic exceeding capacity, current systems
%\todo{can we have some citations?}
employ limited and often ineffective overload management techniques. 
\begin{enumerate}
    \item Rate Limiting: These mechanisms simply reject excess requests without considering their relative importance or potential impact.
    \item Short Request Prioritization: These techniques favor shorter requests, which can unfairly disadvantage longer but potentially more important queries.
\end{enumerate}
Such approaches are unable to provide application-aware or graceful service degradation, resulting in either uniform performance degradation across workloads or complete rejection of a class of requests without any fairness guarantees.

% challenges with current deployments

%1. operational complexity due to varying load between each silo - capacity provisioning is challenging - overall resource underutilization
%If there are applications with different latency requirement, create more silos - adds to complexity

%2. No graceful degradataion. Discuss the downsides of existing overload management techniques.
\subsection{Deployment Challenges}
Current LLM deployments create significant operational inefficiencies due to the siloed infrastructure model.

\jheading{Resource provisioning and utilization} As workload demands fluctuate, dedicated clusters often operate well below their maximum capacity,  resulting in substantial resource underutilization. An interactive cluster might be overwhelmed during peak hours, while a batch processing fleet remains largely idle, leading to inefficient computational resource allocation.
The complexity intensifies when supporting applications with multiple different latency requirements. Each unique performance profile potentially necessitates a dedicated infrastructure cluster, which can increase operational complexity significantly. What begins as a straightforward architectural decision quickly transforms into a management challenge, with each new cluster introducing additional capacity provisioning challenges and monitoring overhead. 

\jheading{Lack of graceful service degradation} Existing mechanisms for overload management, such as user rate limiting and prioritizing short requests, are often unfair and not application-aware, and thus lack an ability to gracefully degrade QoS. These techniques can lead to poor user experiences and inefficient resource utilization.

%analysis of existing scheduling algorithms for multi-SLA scheduling
\subsection{Analysis of Multi-SLA scheduling policies for LLM Inference}
A practical approach to mitigating the operational complexities and resource inefficiencies of siloed infrastructure is to co-schedule requests from various applications within a unified cluster. In this section, we examine the effects of traditional scheduling policies from the literature on multi-tenant scheduling and assess their performance for LLM inference across three key dimensions: latency, SLO violations, and the fairness of SLO violations. This analysis highlights the necessity for a novel multi-tenant, SLO-aware scheduling policy tailored for LLM inference.

\jheading{Scheduling policies} We compare four different scheduling policies from the literature for multi-tenant systems. First-Come-First-Served (FCFS) represents the most basic approach, processing requests in the order they arrive. More advanced policies include Shortest Job First (SJF), which prioritizes jobs with the shortest expected execution time, and Shortest Remaining Prompt First (SRPF), which continuously re-evaluates and preempts jobs to minimize overall waiting time, based on the outstanding prompt tokens to be processed. Finally, Earliest Deadline First (EDF) schedules jobs based on their impending deadlines.
%while Least Slack First (LSF) dynamically schedules jobs based on the remaining time before their deadline. 

\Cref{fig:motivation_sched} compares the multiple scheduling policies and plots the (a) median and (b) p99 latency of requests in the system, (c) percentage of requests that violated their SLO, and (d) the number of long requests (requests with prompt length in the 90th percentile of the dataset) that violated their SLO. Despite their theoretical foundations, we observe that these scheduling approaches fundamentally struggle when applied to large language model (LLM) inference workloads. \sysname exploits the unique computational characteristics of LLMs --- including variable input complexity, distinct prefill and decode phases, and the predictability of the prefill phase to devise an SLO-aware scheduling policy which minimizes latency and SLO violations while maximizing throughput, as we show in our evaluations (\Cref{fig:eval:latency,fig:eval:violations}).

%Traditional scheduling strategies which are oblivious to the computational pattern of LLM inference are inadequate to maximize throughput while minimizing latency and SLO violations.
%Our comprehensive evaluation reveals profound limitations across three critical dimensions: serving latency, deadline violations, and fairness metrics.

% Talk about the eval
%\jm{Insert a graph comparing existing scheduling policies with ours}
%Traditional policies fail to capture the intricate interplay between different computational stages, application-specific performance requirements, and dynamic workload characteristics.

% Summary

%We need a system that addresses both these practical deployment challenges -operationally simple to provision and deploy, but also provide graceful service degrdataion.

\jheading{Summary}
In this paper, we address these critical infrastructure challenges by introducing a QoS-aware serving framework, \sysname.  Our system transforms LLM inference serving from a static, siloed approach to a dynamic, application-aware computational system. By introducing sophisticated service level objective (SLO) management, \sysname enables more efficient, responsive, and cost-effective infrastructure for next-generation AI applications.

%\todo{suggestion: Should we remove the key features below? It has been covered in intro and will be covered again in the design section.}

%\input{sections/4-design-v3}
\section{\sysname: Design and Implementation}
\label{sec:design-impl}
\sysname is designed to efficiently manage concurrent LLM inference requests with diverse QoS requirements, while maximizing resource utilization across the shared infrastructure. We address the limitations outlined earlier by dynamically adapting scheduling decisions based on real-time system state and QoS targets of the in-flight requests. %~\cref{fig:system_design} illustrates the high-level components and interactions within our system. 

\subsection{Overview}
\begin{figure}[t]
    \centering
\includegraphics[width=\linewidth]{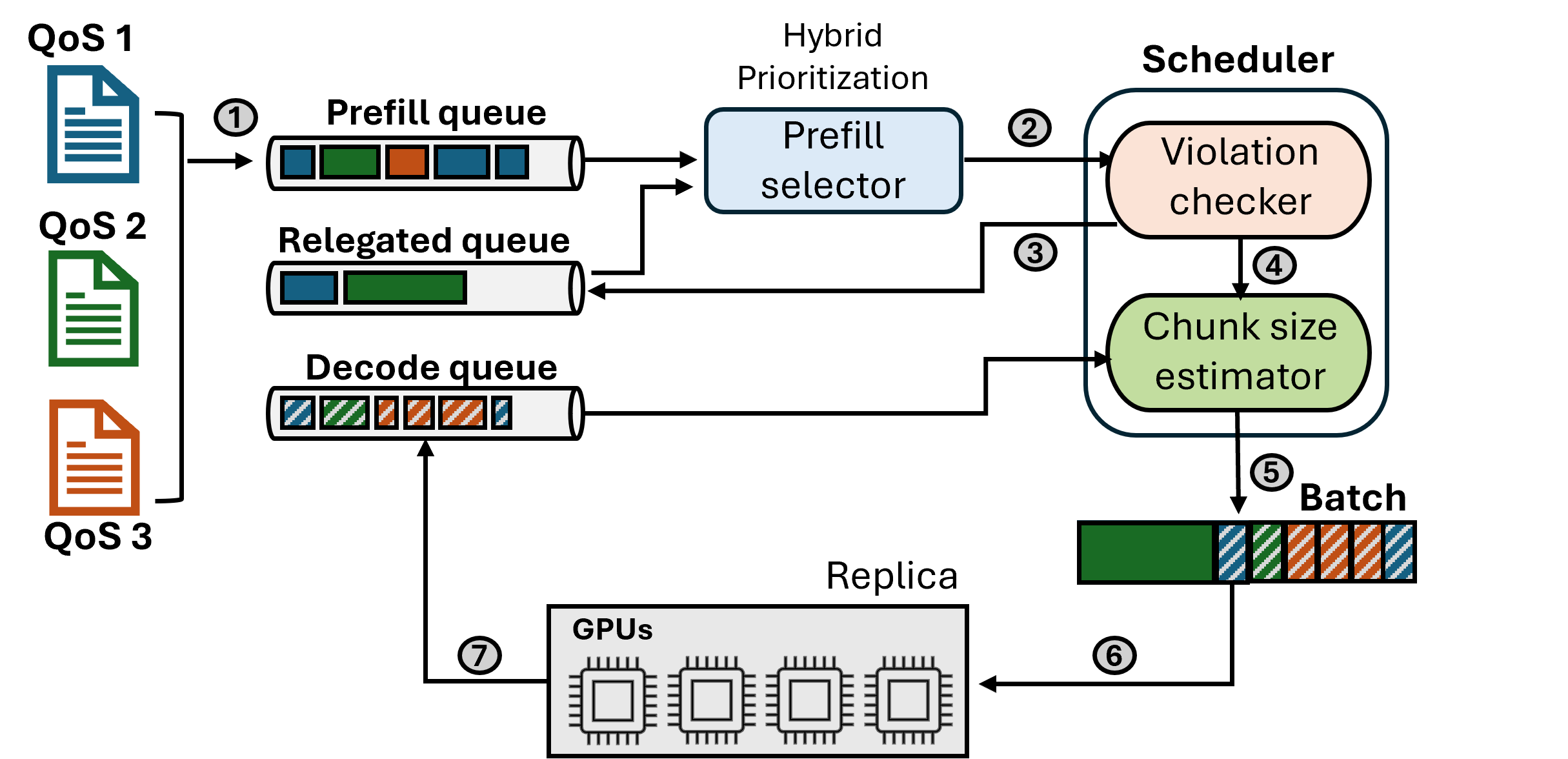}
    \caption{Overview of \sysname}
    \label{fig:system_design}
\end{figure}

The architecture of \sysname is shown in ~\Cref{fig:system_design}. A request in \sysname can be in one of three queues --- 1) prefill queue, 2) decode queue, or 3) relegated queue. \quad \circled{1} When a request enters the system, it is put into the prefill queue. In each iteration, \sysname constructs a batch consisting of all requests in the decode queue, and a prefill-chunk from a request in the prefill queue. The prefill selector uses \emph{hybrid prioritization} to select the prefill request for the current batch. \circled{2} The violation checker module validates that the chosen request has not already (or will) violate its QoS targets in the current iteration. \circled{3} If it does, it is  eagerly moved into the relegated queue and a different prefill request is chosen. The relegated requests are serviced opportunistically during periods of lower system load, ensuring eventual completion without permanent rejection; while enabling graceful degradation under overload conditions. \circled{4} A lightweight predictor is then used to estimate the latency of the batch to make sure that the QoS targets are not violated, while maximizing the chunk size for efficiency. \circled{5} A mixed batch of prefill and decode tokens is constructed using the chosen prefill chunk and the requests in the decode queue, which is then \circled{6} dispatched to the execution engine on the GPU for processing. \circled{7} Once the prefill portion of a request is completed, it is moved to the decode queue, and subsequent iterations continue.

%The prefill request and the corresponding chunk size is chosen based on the QoS targets of the requests in the system and tenant priorities. A lightweight predictor is used to estimate the latency of the batch to make sure that the QoS targets are not violated, while maximizing the chunk size for efficiency. If at any time the system realizes that the QoS target of a request in the prefill queue cannot be met at the existing load, the request is eagerly moved into the relegated queue. The relegated requests are serviced opportunistically during periods of lower system load, ensuring eventual completion without permanent rejection; while enabling graceful degradation under overload conditions. Once the prefill portion of a request is completed, it is moved to the decode queue. \todo{add psuedo code}

\label{sec:design}

\subsection{QoS Classes and Deadlines}
\sysname defines two QoS classes: interactive and non-interactive. Interactive requests use two SLOs --- TTFT (time to first token) and TBT (token-by-token latency), which ensure immediate responsiveness and consistent pacing. Non-interactive requests have a single TTLT (total latency) target, focused on overall completion. Although, we define two QoS classes, the application owner is free to specify their custom SLO targets within the class, allowing for flexibility and customization to specific application needs as shown in ~\Cref{tbl:eval:qos}.

The deadline for each request is determined based on its QoS class. For the interactive QoS class, following the approach in ~\cite{etalon2024}, the deadline for the first token is defined as:
\begin{equation}
\label{eq:deadline_ttft}
    D_{first} = t_{arrival} + SLO_{TTFT},
\end{equation}
while subsequent tokens' deadlines are calculated using:
\begin{equation}
\label{eq:deadline_tbt}
    D_n = t_{arrival} + SLO_{TTFT} + (n - 1) \cdot SLO_{TBT},
\end{equation}
where $n$ is the token position. For non-interactive requests a deadline is set only for the full completion of the request as:
\begin{equation}
\label{eq:deadline_ttlt}
    D_{total} = t_{arrival} + SLO_{TTLT}
\end{equation}

Once we have defined the deadlines for each request, \sysname scheduling aims to minimize deadline violations while maximizing throughput.

\subsection{Dynamic Chunking}

\begin{figure}[t]
    \centering
    \includegraphics[width=0.9\linewidth]{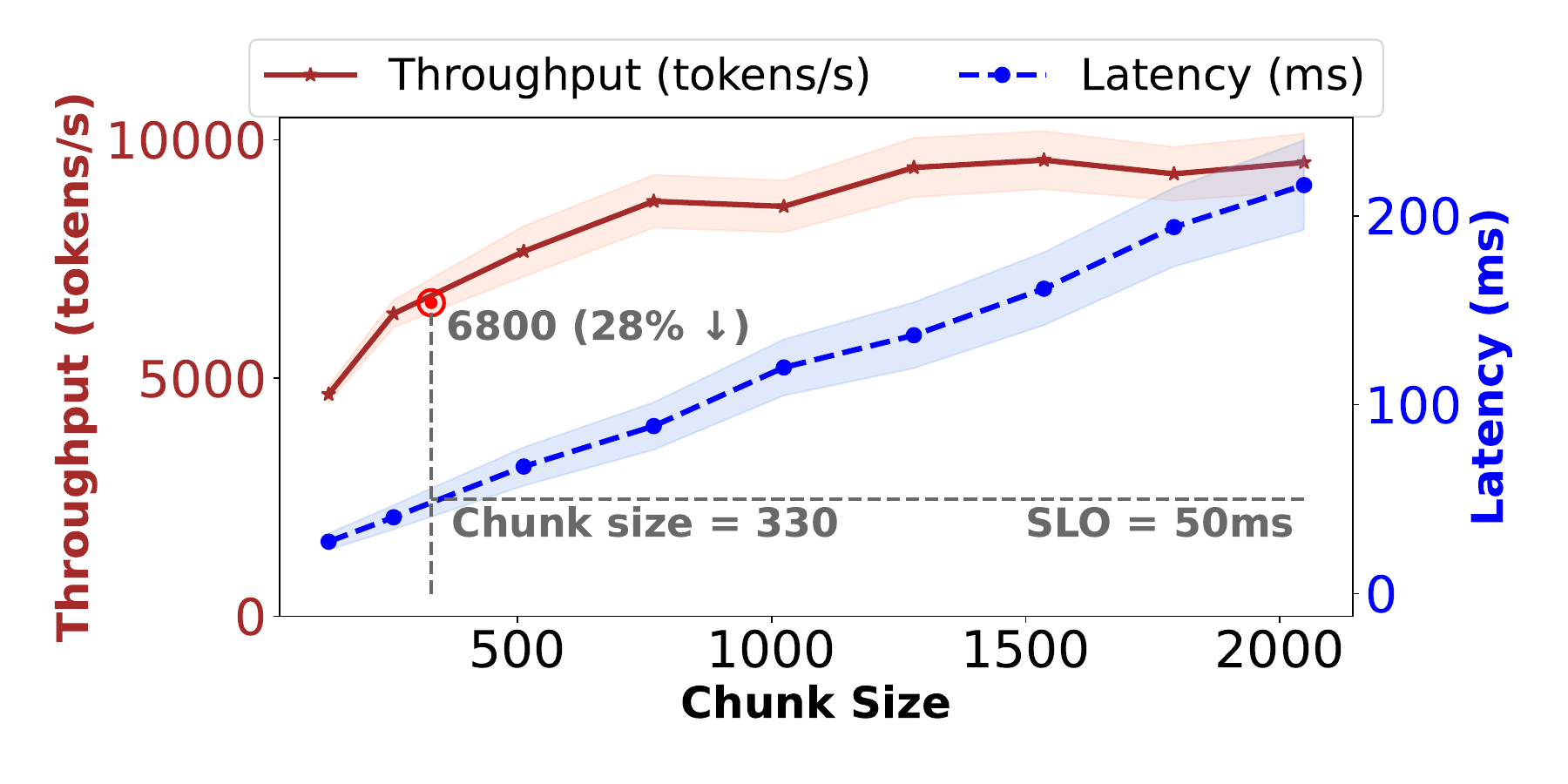}
    \caption{Performance characteristics as a function of chunk size, showing the throughput-latency tradeoff.}
    \label{fig:chunk-perf}
\end{figure}

%LLM inference scheduling involves a fundamental trade-off between throughput and latency~\cite{agrawal2024taming}.
%, i.e.,  reducing latency requires splitting user input into smaller chunks but this results in lower throughput\footnote{An alternative approach is using a disaggregated architecture~\cite{patel2023splitwise,distserve2024} but that comes with its own trade-offs.}.
State-of-the-art LLM inference serving frameworks~\cite{vllmsosp, sarathi2024} serve requests using chunked-prefills, where each iteration processes a fixed number of tokens (called chunk size), which includes both prefill and decode tokens from different requests using fused prefill-decode MLP to improve the compute efficiency of memory-bound decode phase~\cite{sarathi2023}. However, this involves a fundamental trade-off between throughput and latency -- a larger chunk results in better throughput but increases the TBT of the decodes in the batch. This is illustrated in ~\Cref{fig:chunk-perf}.
%Therefore, interactive jobs use smaller chunk sizes (served at higher cost per token due to lower throughput) while batch jobs use larger chunk sizes (served at lower cost thanks to higher throughput).

A na\"ive approach for co-scheduling jobs of different QoS classes with varying deadlines on TTFT, TBT, and TTLT would be to use the smallest chunk size necessary to meet the latency constraint of the strictest QoS class. However, this results in low throughput and high cost for all service classes. 

\sysname employs \textit{dynamic chunking} to opportunistically maximize the chunk size for the prefill request by exploiting any available slack in the deadlines of the requests being currently serviced. For each request in the decode queue, we define slack as the difference between the deadline for the next token (eq~\ref{eq:deadline_tbt}) and current time. %To estimate the time it will take to process a given batch (which will include all decode requests and a prefill chunk), we build a lightweight predictor using offline profiles from an LLM simulator~\cite{vidur} (more details in section~\ref{sec:batch_predictor}\todo{add a section for predictor ore remove this reference}).
Using this slack and characteristics of the requests in decode phase, we calculate the maximum chunk size which can be used for the current batch without violating any deadline.

%\sysname opportunistically increases chunk size in every iteration based on the accumulated slack(time remaining before their respective deadlines) of all the requests in the current batch. Our system dynamically adjusts chunk size based on accumulated slack (time before deadlines) of all requests in the current batch. This adaptive approach responds to real-time conditions: when requests have sufficient slack, larger chunks maximize throughput; as deadlines approach, chunk sizes decrease to ensure timely processing. We leverage the predictability of LLM inference latency given its batch composition by building a lightweight predictor using offline profiles from an LLM simulator~\cite{vidur}. Building a chunk size predictor for a model-hardware combination requires only about an hour of profiling on a single target GPU. 

\subsection{\sysname Scheduling}
\label{sec:sched_optimizations}
While dynamic chunking allows us to choose an optimal chunk size for a prefill request, we also need to decide which request from the prefill queue should be processed in the current scheduling iteration.

\jheading{Hybrid Prioritization}
As shown in ~\Cref{fig:motivation_sched}, existing scheduling policies struggle with LLM workloads at higher loads. For example, EDF which prioritizes requests with earlier deadline has very low deadline violation rates (\Cref{fig:motivation_sched}(c)) at low loads, but the violation rates spike to almost 100\% when the load goes beyond a threshold. On the other hand, policies which prioritize short work requests --- SRPF and SJF --- handle higher loads much better but are worse than EDF at lower loads. Further, SRPF and SJF achieve this at the expense of unfairly penalizing long jobs (\Cref{fig:motivation_sched}(d)) without any regard to the request priorities while EDF does not suffer from this problem. To handle varying load conditions which is common in production services and maintain fairness across requests, our first key insight is a \textit{hybrid prioritization} scheme which interpolates between SRPF and EDF. This allows us to get EDF characteristics at low loads, and leverage SRPF semantics under overload conditions while maintaining fairness.

To implement this scheduling, \sysname smoothly interpolates between EDF and SRPF to compute the priority of a request. For interactive requests, the priority is computed by taking a linear combination of the TTFT deadline (this incorporates EDF semantics) and the estimated time taken which will be needed to process the remaining prefills (this incorporates SRPF semantics) of the request as:
\begin{equation}
\label{eq:priority_interactive}
    P^i = t_{arrival}^i + SLO_{TTFT}^i + \alpha * Prefill_{rem}^i.
\end{equation}
Note that we only consider TTFT deadline as meeting TBT deadlines is taken care by our dynamic chunking scheme. For non-interactive requests, the priority is computed as
\begin{equation}
\label{eq:priority_batch}
    P^i = t_{arrival}^i + SLO_{TTLT}^i + \alpha * (Prefill_{rem}^i + Decode_{rem}^i),
\end{equation}
where $Decode_{rem}$ indicates the time to compute all the decode tokens. Since decode length is unknown in LLM inference, this introduces a challenge in modeling the priority of non-interactive requests. We address this with a simple insight --- for non-interactive jobs, the TTLT deadline is typically much greater than the actual processing time. Therefore, given an application, we can use historic information on the decode tokens generated by that application and over-approximate it by two standard deviations. We show in \sref{sec:eval} that this simple prediction sufficiently captures the priority of non-interactive jobs.
 
\begin{figure}[t]
    \centering
\includegraphics[width=0.95\linewidth]{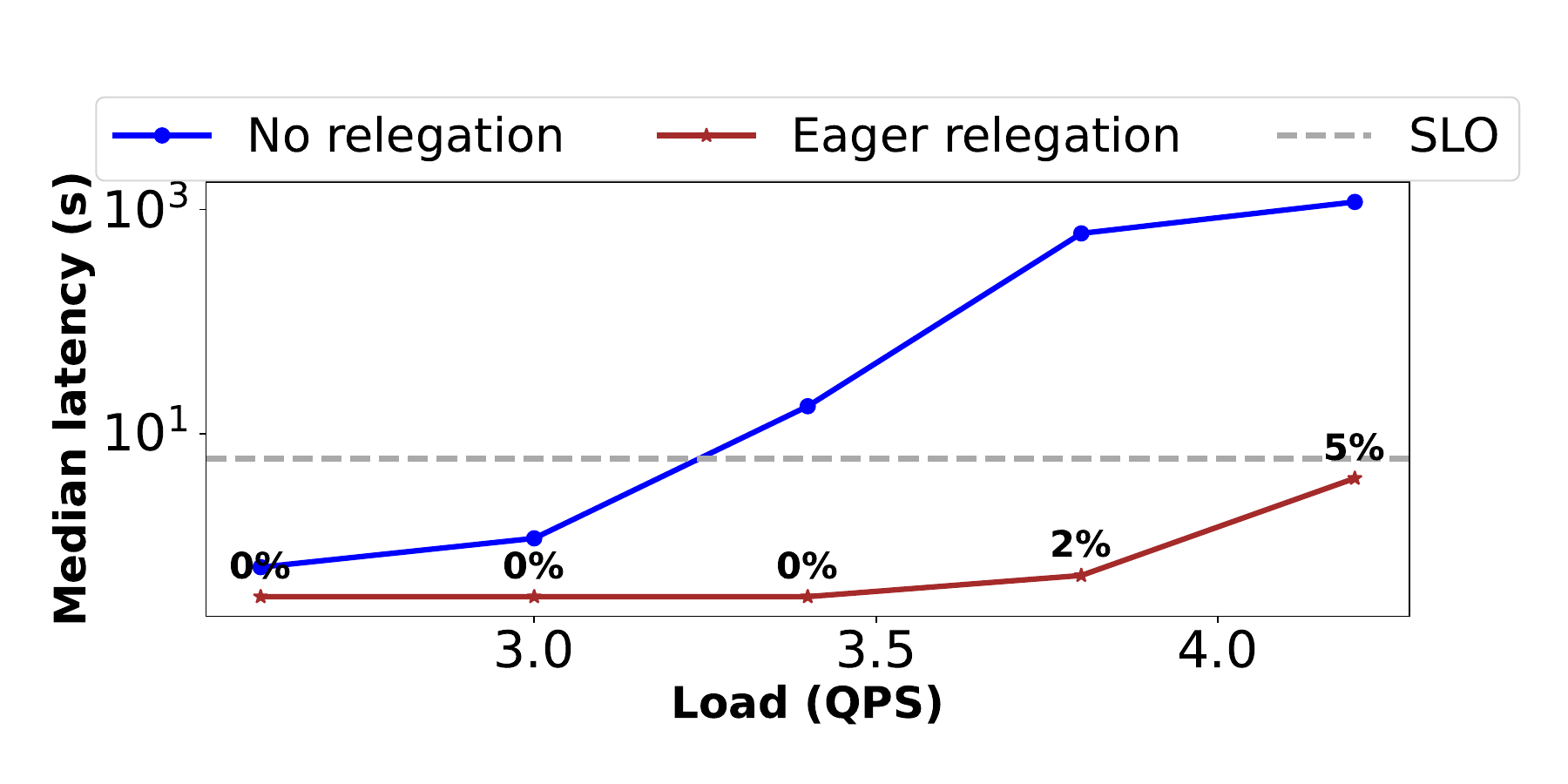}
    \caption{Proactively relegating a small percentage of requests enormously helps is maintaining the quality of service for the median request in the system, which otherwise grows exponentially due to a cascade of violations.}
    \label{fig:motivation_relegation}
\end{figure}

\jheading{Eager Relegation}
Our \textit{hybrid prioritization} scheduling strikes a balance between minimizing deadline violations and fairness. However, as shown in~\Cref{fig:motivation_relegation}, under overload conditions, \sysname (or any scheduling policy) still cannot service all incoming requests at the desired QoS SLOs. Our second key insight is that by \textit{eagerly relegating} a small fraction of requests that we know will miss their deadlines, one can provide stable performance for the majority, enabling graceful service degradation under overload conditions. The key idea is simple --- if a request has already violated its TTFT / TTLT deadline, or is about to violate it in the current iteration, then \sysname de-prioritizes this request into a relegated queue. In multi-tenant deployments, we also use application hints such as free vs paid tier to preferentially relegate low-priority requests to ensure stability of service to the high priority ones. Only when there are no more low-priority requests, \sysname proactively relegates high-priority requests that have violated their deadlines to prevent cascading deadline violations. This enables graceful degradation of service even under extreme load. As shown in ~\Cref{fig:motivation_relegation}, by relegating just 5\% of the requests, we can maintain latency SLOs even under very high overload conditions.

\jheading{Selective Preemption} Note that our \textit{hybrid prioritization} scheduling can preempt an in-flight request for which a few prefill chunks have already been processed to instead service a new incoming request with a strict QoS target with a higher computed priority (eq~\ref{eq:priority_interactive}). Preemption is a desirable capability as it avoids head-of-line blocking of small interactive requests behind long batch requests. However, in LLM serving, the memory overhead of preemption can be significant as the KV-cache of requests can be quite large.
%It is possible that a new incoming request with a strict QoS target gets a higher priority (eq~\ref{eq:priority_interactive}) compared to an in-flight request for which a few prefill chunks have already been processed. A priority based system will preempt processing of this in-flight request in future iterations, and instead schedule chunks from the new request.
%However, this can result in a blow up in the KV-cache as one will need to hold onto the KV-cache of these preempted request or pay the cost of recomputing them later.
To avoid this, \sysname uses \textit{selective preemption}, where we preempt a request to accommodate another with a higher priority only if (1) the in-flight request is in the prefill queue (i.e., requests in the decode queue are never preempted), and (2) preempting that request for an iteration does not result in its deadline violation. We do not preempt requests in the decode queue as TBT latency targets are typically very strict (10s of milliseconds), and thus preempting them significantly increases the chances of TBT violation. This approach also ensures that the KV-cache for each request remains in the GPU for the shortest necessary duration, thereby minimizing memory pressure. 
%todo{the kv-cache argument is not very convincing)}.

\subsection{An Illustrative Example}
\begin{figure}[t]
    \centering
\includegraphics[width=\linewidth]{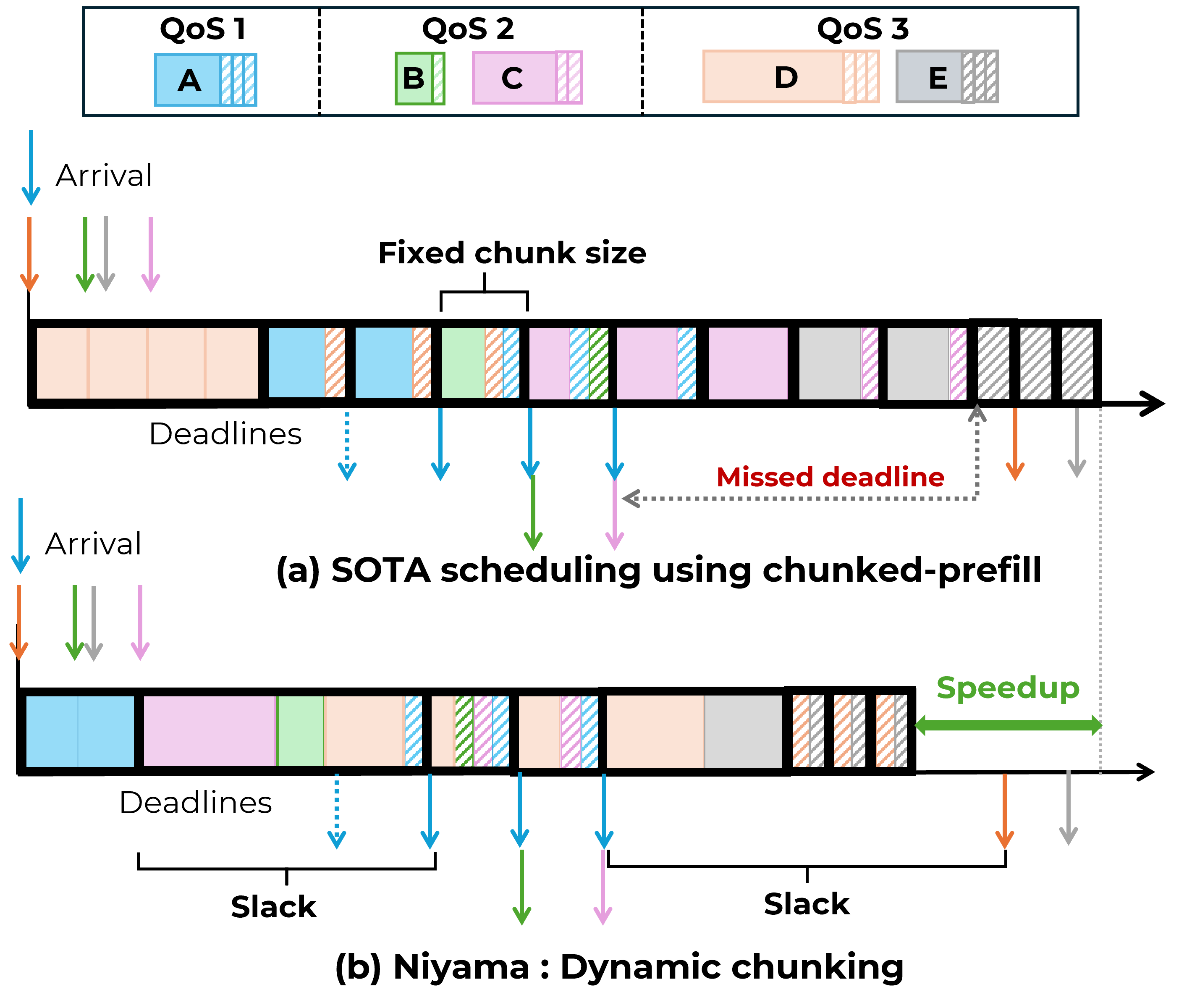}
\vspace{-2em}
    \caption{An illustration of how \sysname improves throughput using dynamic chunking compared to SOTA scheduling.}
    \label{fig:dynamic_chunk}
\end{figure}
\Cref{fig:dynamic_chunk} illustrates our approach with an example of five requests (A--E) across three QoS buckets. In this example, A is an interactive request while others are non-interactive. State-of-the-art LLM schedulers like vllm~\cite{vllmsosp} and sarathi~\cite{sarathi2024} will execute each iteration using a fixed chunk size and process requests in arrival order (FCFS). Our solution introduces two key improvements.

First, we prioritize requests based on their QoS targets using our hybrid prioritization, which will prioritize request A before D due to its earlier deadline.
%\sref{sec:sched_optimizations} details how we further adapt this traditional policy to address challenges and opportunities unique to LLM inference.
Second, we dynamically adjust chunk sizes based on accumulated slack. For example, after A's prefill phase completes earlier than its deadline, it accumulates significant slack before its next token is due. We exploit this slack by dynamically increasing the chunk size, adding more prefill tokens from requests B and D (which have the earliest deadlines in the queue), thereby improving throughput without violating any ongoing request deadlines.

When the interactive job A enters its decode phase, we revert to the original smaller chunk size necessary to meet its TBT requirement, though we still exploit any slack accumulated if decoding completes faster than predicted. Once A completes and no remaining requests impose strict TBT constraints, we again increase chunk size to maximize throughput while respecting the TTLT deadlines of ongoing requests. This approach effectively leverages the predictable execution characteristics of LLMs to dynamically optimize chunk sizes during runtime, balancing throughput and deadline requirements.

\subsection{Implementation}

We implemented \sysname by extending the Sarathi scheduler~\cite{sarathi2024}, which is built on top of the vLLM inference system~\cite{vllmsosp}. The implementation focuses on enhancing the scheduler component while maintaining compatibility with vLLM's efficient tensor parallelism and PagedAttention mechanisms. We extended the vLLM API to associate each inference request with its corresponding QoS requirements (TTFT, TBT, and/or TTLT) and priority level during request submission. The dynamic chunk size predictor analyzes the characteristics of requests during the decode phase and determines an optimal chunk size that maximizes throughput while adhering to latency constraints. It is implemented as a lightweight random forest model, trained on performance profiles collected from the Vidur LLM inference simulator~\cite{vidur}, ensuring minimal overhead in the scheduling process. The hybrid prioritization policy is implemented using a priority queue that incorporates both deadline proximity and estimated processing time, with the interpolation factor $\alpha$ configurable as a deployment parameter. For non-interactive requests, we maintain a running history of token generation patterns per application to estimate the expected decode length.
%The eager relegation mechanism tracks the running throughput of the system, estimates the expected processing time of each request, and determines whether to relegate a request based on these metrics. Relegated requests remain within the same scheduler, residing in a lower-priority queue that may be serviced during periods of lower system load.
To support multi-tenant deployments, we added a priority field to each request that enables informed relegation decisions based on application hints such as free-tier versus premium users.
\section{Evaluation}
\label{sec:evaluation}

\begin{comment}
    \begin{table}[t!]
    \centering
    \scalebox{0.85}{
    \begin{tabular}{l|c|}
     Model & GPU  \\ \toprule
     Llama3-8B & 1 A100 \\
     Llama3-70B & 4 A100s \\ \bottomrule
    \end{tabular}}
    \caption{Models and hardware used for evaluation.}
    \label{tbl:eval:models}
\end{table}
\end{comment}

\begin{table}[t!]
    \centering
    \scalebox{0.85}{
    \begin{tabular}{c|cc|cc}
     \multirow{2}{*}{Dataset} & \multicolumn{2}{c|}{Prompt tokens} & \multicolumn{2}{c}{Decode tokens} \\ 
     & p50 & p90 & p50 & p90 \\ \toprule
     ShareGPT & 1730 & 5696 & 415 & 834 \\
     Azure Conv &  928 & 3830 & 41 & 342 \\ 
     Azure Code & 1930 & 6251 & 8 & 43 \\ \bottomrule
    \end{tabular}}
    \caption{Datasets used in evaluation}
    \vspace{-2em}
    \label{tbl:eval:datasets}
\end{table}

Our evaluation aims to answer the following questions.
\begin{enumerate}[noitemsep]
    \item What is the improvement due to \sysname in the serving capacity while meeting specified QoS SLOs~\sref{sec:eval:tput}.
    \item What is the impact of \sysname on request latencies and deadline violations under high load conditions~\sref{sec:eval:latency}. 
    \item How does \sysname respond to transient spikes in load~\sref{sec:eval:overload}.
    %\item Does \sysname enable higher serving throughput while maintaining deadline violations below a threshold ~\sref{sec:eval:tput} and improve latency and SLO violations of requests with varying load~\sref{sec:eval:latency}?
    %\item Does \sysname gracefully adapt to overload conditions~\sref{sec:eval:overload}?
    \item What is the independent impact of the different optimizations and design choices used in \sysname~\sref{sec:eval:ablation}?
\end{enumerate}

\jheading{Models and Hardware} We evaluate our system across two different models and parallelism stategies; Llama3-8B model deployed on a single A100 GPU with 80GB physical memory, and Qwen-7B depoyed across two A100 GPUs (TP2). We evaluate them across three different popular open-source datasets which have different ratios of prefill to decode tokens, as shown in \Cref{tbl:eval:datasets}. %We focus on dataset heterogeneity rather than model heterogeneity as scheduling algorithms are more sensitive to the dataset than the model. We have validated that our observations hold across different models as well as tensor-parallel deployments.

\jheading{Workloads and QoS Tiers}
%We focus on evaluating serving capacity and latency in an online inference scenario.
For workloads, we use popular open-source datasets such as ShareGPT~\cite{sharegpt} and coding and conversation production traces from multiple LLM inference services in Azure~\cite{dynamollm}. Request arrival times are generated using a Poisson distribution, ~\cite{sarathi2024,prabhu2025vattentiondynamicmemorymanagement}, while maintaining the prefill and decode token counts of the respective traces. To emulate different applications, we divide each dataset into three equal parts, and assign each part with a different application type and the corresponding QoS bucket and SLO. We consider three QoS buckets: one interactive and two non-interactive, as shown in \Cref{tbl:eval:qos}.

\begin{table}[t!]
    \centering
    \scalebox{0.85}{
    \begin{tabular}{c|c|cc|c}
     QoS &  Request & \multicolumn{2}{c|}{Interactive} & Non-interactive \\ 
     bucket & ratio & TTFT(s) & TBT(ms) & TTLT(s) \\ \toprule
     Q1 & 33.33\% & 6 & 50 & - \\
     Q2 & 33.33\% & - & - & 600 \\ 
     Q3 & 33.33\% & - & - & 1800 \\ \bottomrule
    \end{tabular}}
    \caption{QoS classes and workload composition}
    \label{tbl:eval:qos}
    \vspace{-2em}
\end{table}
\begin{figure*}[!t]
    \centering
    \begin{subfigure}[t]{0.45\textwidth}
         \includegraphics[width=\linewidth]{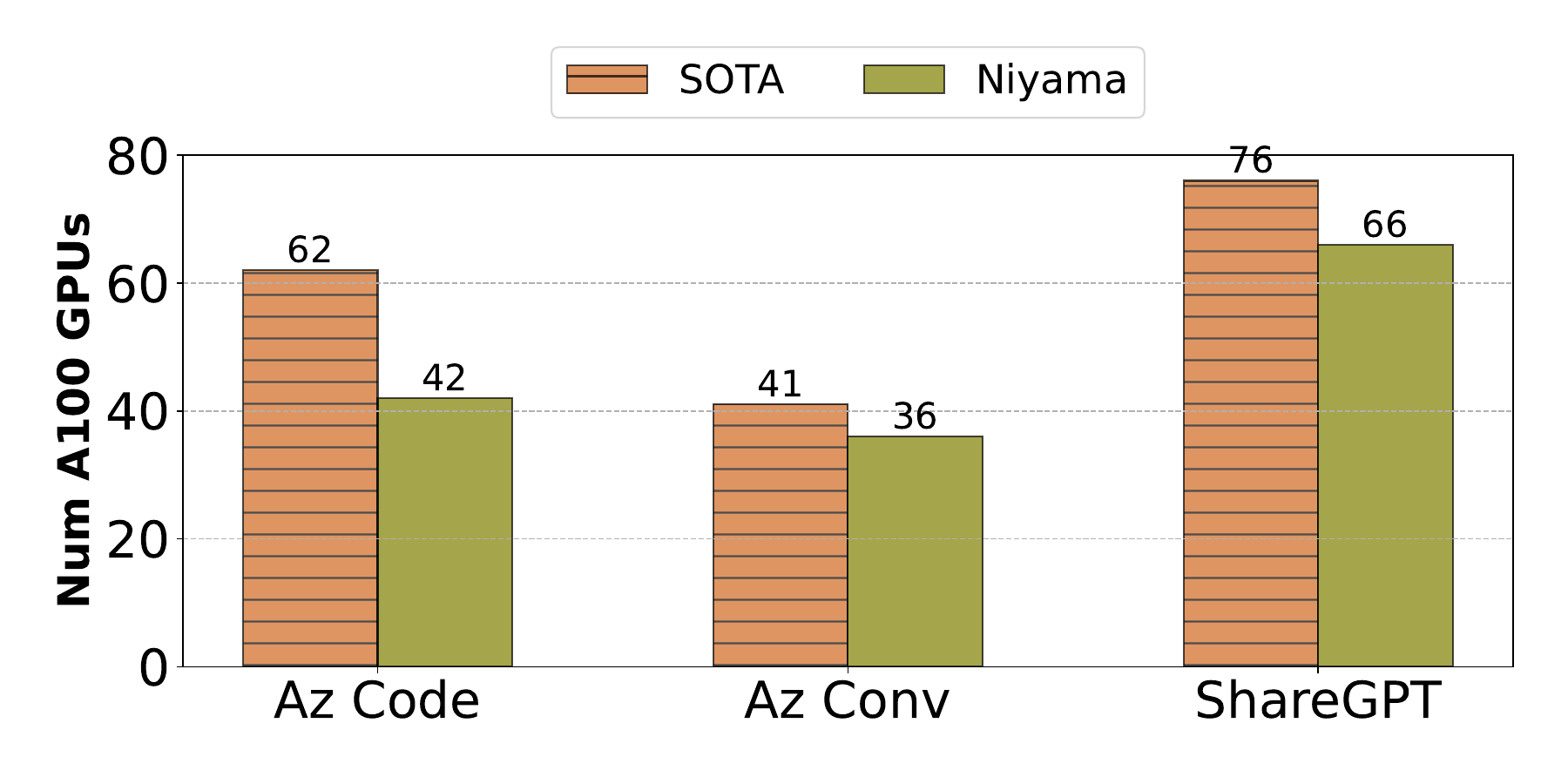}
         \caption{Capacity to serve a load of 50 QPS}
         \label{fig:eval:gpus}
     \end{subfigure}
     \begin{subfigure}[t]{0.45\textwidth}
         \includegraphics[width=\linewidth]{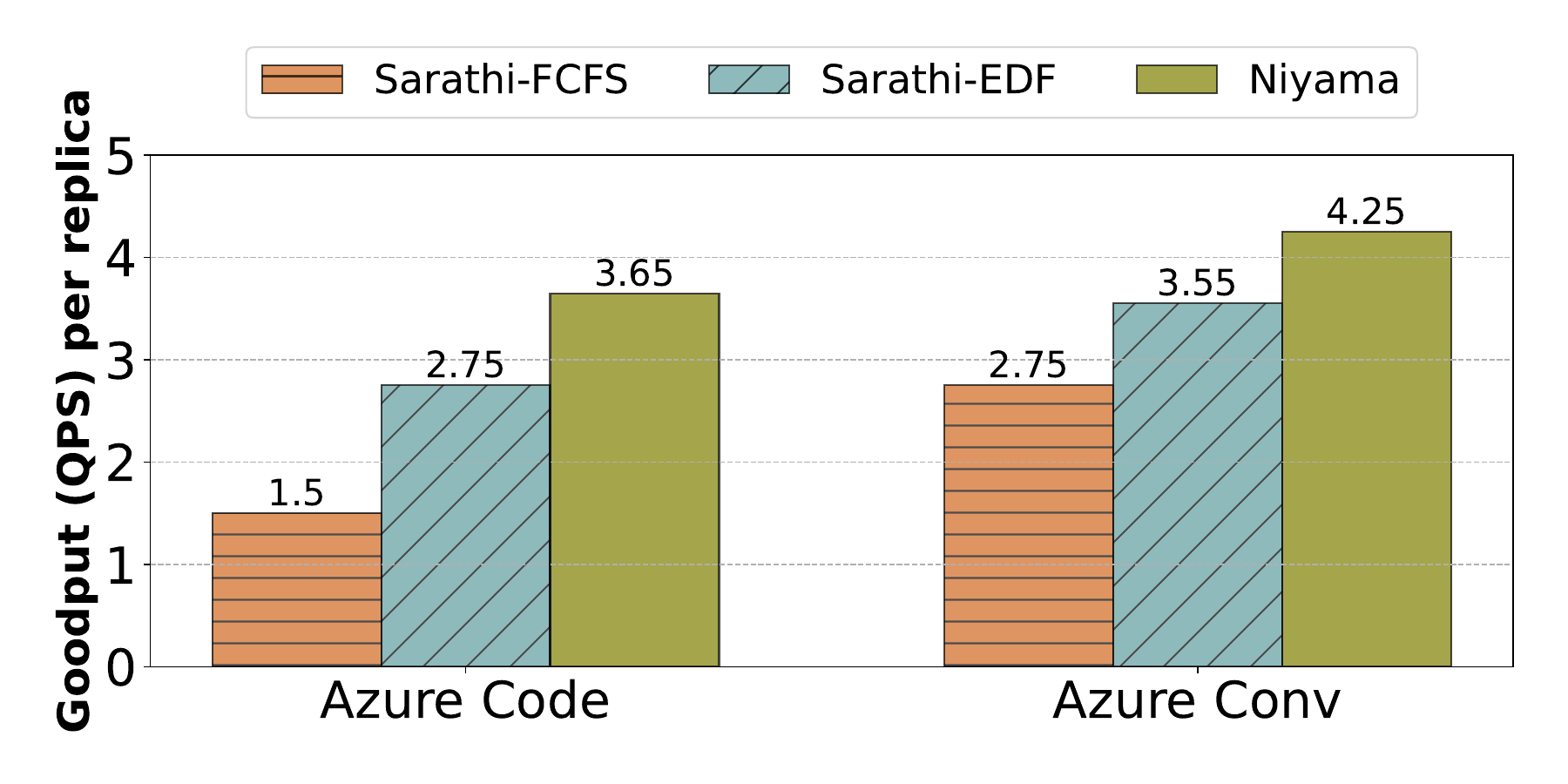}
         \caption{Maximum goodput in a shared cluster}
         \label{fig:eval:goodput}
     \end{subfigure}
     \vspace{-1em}
        \caption{Capacity evaluation at uniform load}
        \vspace{1em}
\end{figure*}

\begin{figure*}[t]
    \centering
    \includegraphics[width=0.95\linewidth]{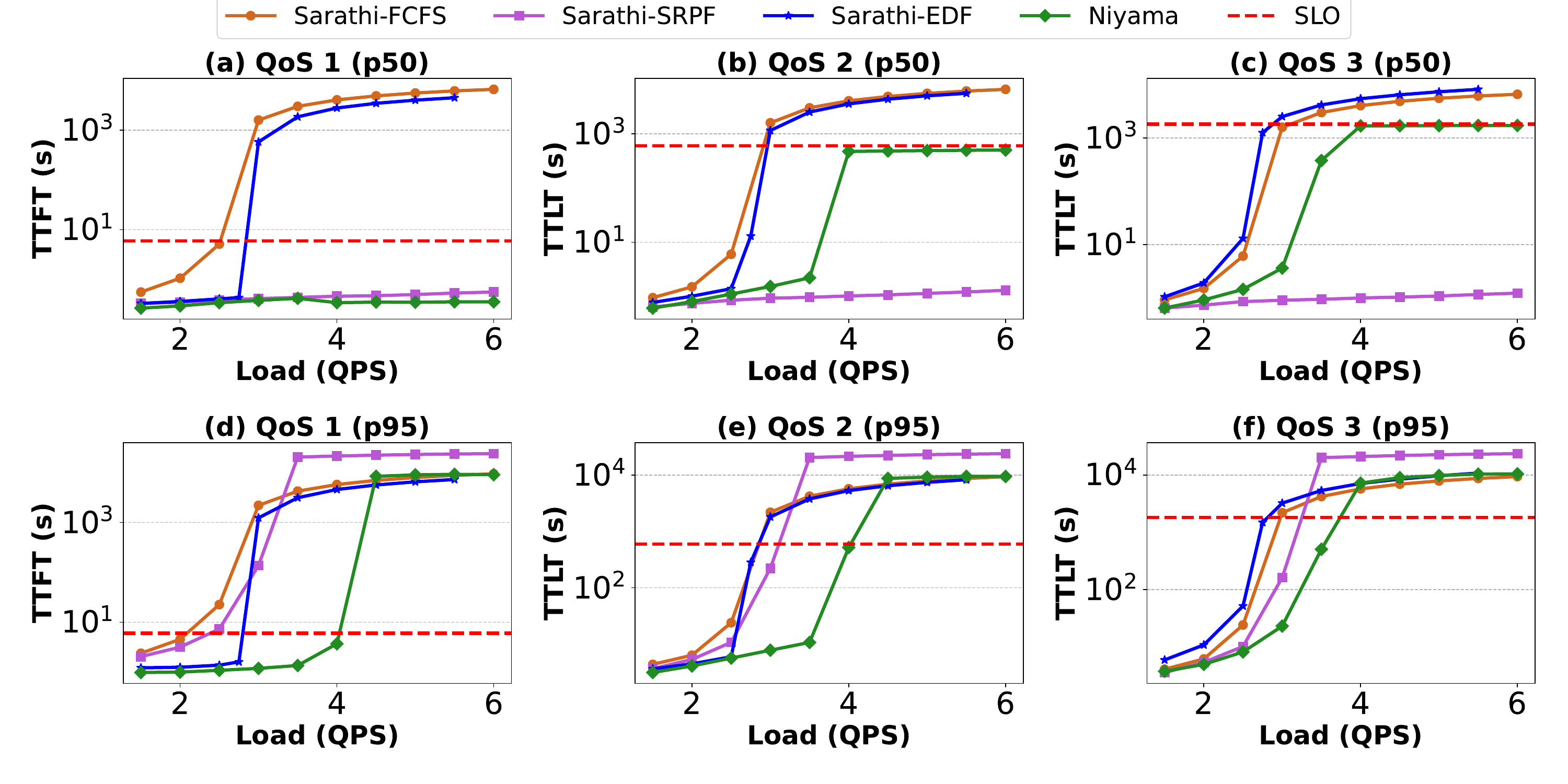}
    \vspace{-1em}
    \caption{Latency of requests across the three QoS buckets as we vary load in the system}
    \label{fig:eval:latency}
\end{figure*}

    % \begin{subfigure}[b]{0.95\linewidth}
    %     \includegraphics[width=\linewidth]{graphs/llama3_8b_latency.pdf}
    %     \caption{Median latency}
    % \end{subfigure}

    % \begin{subfigure}[b]{0.95\linewidth}
    %     \includegraphics[width=\linewidth]{graphs/llama3_8b_latency_p95.pdf}
    %     \caption{p95 latency}
    % \end{subfigure}

\jheading{Baselines} We built \sysname on top of Sarathi-Serve~\cite{sarathi2024}, which itself extends vLLM~\cite{vllmsosp}. Our evaluation includes several baseline configurations: (1) \textbf{Sarathi-Silo (SOTA)}, the state-of-the-art siloed deployment where each QoS bucket is assigned an independent GPU cluster with each replica running a Sarathi scheduler (2) \textbf{Sarathi-FCFS}, which co-schedules requests across all QoS Tiers on a unified cluster using Sarathi with FCFS prioritization policy, and (3) \textbf{Sarathi-EDF}, which again co-schedules but also imparts deadline-awareness during scheduling by using the Earliest Deadline First policy on Sarathi. The strictest QoS bucket with 50ms TBT deadline uses a chunk size of 256, while the other two QoS classes use a large chunk size of 2K to maximize throughput in the siloed baselines. For shared cluster baselines, the chunk size chosen is 256, to meet the TBT targets of the strictest tier.

%This is an optimized baseline that we implement on top of Sarathi to demonstrate the efficacy of the scheduling optimizations in \sysname.
%These baselines allow us to isolate and evaluate the benefits of both resource unification and deadline-aware scheduling before comparing with \sysname's full capabilities.

\jheading{Setup} We first evaluate \sysname under uniform load conditions to identify the impact of our design on goodput~\sref{sec:eval:tput} as well as on latency and SLO violations~\sref{sec:eval:latency}. Next, we evaluate how \sysname performs under transient spikes in load~\sref{sec:eval:overload}, and finally we perform detailed ablation of our individual techniques~\sref{sec:eval:ablation}. %For evaluations on regular load, we assume a poisson arrival distribution of requests given a certain QPS (queries per second).

\subsection{Capacity Evaluation at Uniform Load}
\label{sec:eval:tput}
\subsubsection{Cost Efficiency from Co-located Scheduling}
%We compare the state-of-the-art siloed deployments against \sysname by comparing
We measure the cost benefits due to our co-located scheme compared to the siloed baseline. For this, we compute the number of A100 GPUs required to serve a total load of 50 queries per second (QPS) spread equally across requests from the three QoS classes described in  \Cref{tbl:eval:qos} on three different datasets. We define the serving throughput per replica as the maximum load that can be sustained while violating QoS targets for at most 1\% of the total requests.

\Cref{fig:eval:gpus} shows this computation for the different systems. Across different workloads, \sysname reduces GPU requirements by $13$–$32\%$ compared to the state-of-the-art (SOTA) siloed baseline. This reduction in resources directly translates to cost savings. \sysname achieves better resource efficiency by maximizing throughput while meeting QoS targets through dynamic chunking. Since the lengths of requests (prompt length as well as number of decode tokens) can vary over time, even at uniform QPS the compute load on the serving system varies. \sysname benefits from dynamically increasing chunk size by exploiting any deadline slack of the non-interactive QoS requests as well as any slack of interactive requests during lower load. On the other hand, the siloed replicas serving the strict QoS requests are limited by the small chunk sizes required to meet the TBT constraints resulting in lower efficiency.

%Siloed replicas serving stricter QoS buckets can only handle lower loads because they use smaller chunk sizes to meet the TBT constraints. In contrast, by co-scheduling requests with different deadlines, \sysname can opportunistically maximize throughput by exploiting slack in the deadline across all requests in a batch.

%\input{figures/eval_gpus}
%\input{figures/eval_goodput}

\subsubsection{Goodput}
\label{sec:eval:goodput}
%We now evaluate how \sysname performs compared to existing schedulers in a shared cluster.
We measure the system's goodput, which we define as the number of requests served per second while meeting the latency targets (p99). We allow at most 1\% of total requests to violate their deadlines.
\Cref{fig:eval:goodput} shows the goodput while serving requests over a 4-hour period from the \code dataset on a shared A100 cluster. As shown, \sysname achieves 1.5× to 2.4× higher goodput compared to \fcfs and 20–40\% higher goodput than \edf. These performance benefits stem from a combination of dynamic chunking, hybrid prioritization, and eager relegation. \sref{sec:eval:ablation} examines the contribution of each of these techniques.

%\sysname outperforms traditional scheduling policies using a combination of techniques; dynamic chunking, which balances meeting QoS requirements with maximizing throughput alongside hybrid prioritization and relegation that proactively deprioritizes a small set of requests to maintain QoS for majority of the requests in the system. 

\begin{figure*}[t]
    \centering
\includegraphics[width=0.95\linewidth]{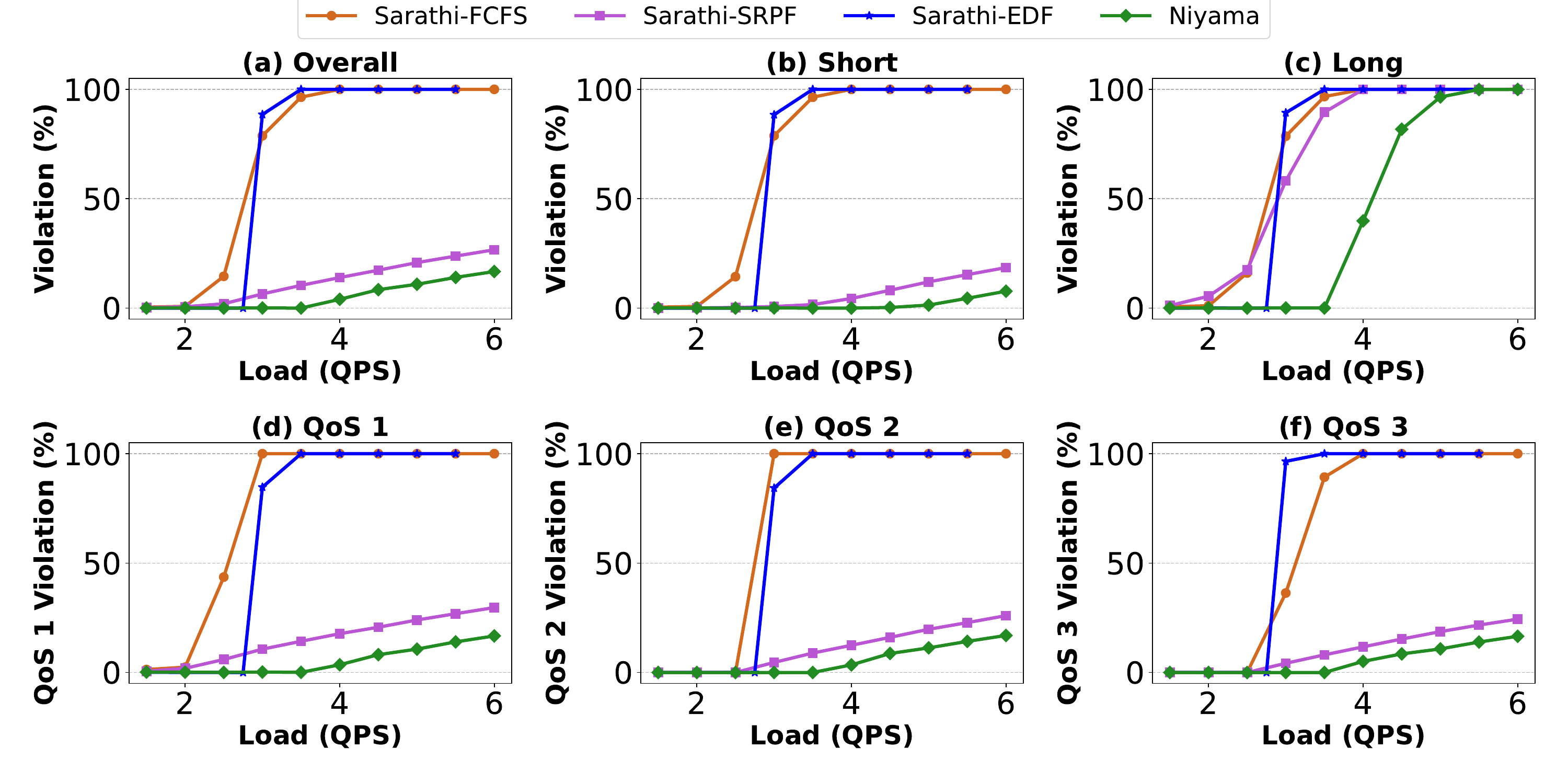}
\vspace{-1em}
    \caption{Deadline violations across all jobs, split by request length and QoS buckets}
    \label{fig:eval:violations}
\end{figure*}

\subsection{Latency and SLO violations under Overload}
\label{sec:eval:latency}
We comprehensively evaluate system behavior at various loads by comparing \sysname against baselines on a shared cluster. We measure three key parameters: (1) median and p95 latency (TTFT, TBT, and TTLT) across all requests, (2) percentage of deadline violations across all SLO buckets, and (3) deadline violations in each SLO bucket and violations categorized by request length to assess scheduling fairness. For this evaluation,  we add another baseline, \srpf which prioritizes jobs with the lowest pending prefill tokens.

\jheading{Latency} \Cref{fig:eval:latency} shows the median and p95 latency across all requests for Llama3-8B on the \code dataset. As load exceeds the optimal operating point, queuing delay increases because the system cannot process requests as fast as they arrive. This causes a sharp increase in latency for all requests. While this happens in every system, the point where scheduling delay becomes unreasonably large defines the maximum serviceable load. We skip showing TBT  in the plots because across all schemes, the average TBT violations was less than 0.1\%, by virtue of carefully chosen chunk size.

\noindent
There are several takeaways from these graphs.
\begin{itemize}[noitemsep, leftmargin=2em, topsep=0pt]
%\begin{itemize}
    \item The state-of-the-art \fcfs scheduler ignores individual request deadlines. As load increases, SLOs for jobs with stricter QoS requirements are violated. At heavy overloads, head-of-line blocking causes denial of service to all requests.
    \item Adding deadline awareness through mechanisms like EDF (\edf) better maintains QoS than \fcfs, but doesn't scale well with load because we must sacrifice throughput to meet SLOs for the strictest QoS tier. It also degenerates at high loads similar to FCFS due to head-of-line blocking. 
    \item Schedulers that prioritize short jobs like \srpf maintain good median latency at the expense of tail latency. The p95 latency, however, grows unbounded because SRPF ignores longer requests. Since it is not deadline-aware it prioritizes minimizing latency across requests of all SLO-buckets, which could have otherwise been used to prioritize those with stringent SLOs.
    \item \sysname handles up to 40\% higher load while meeting tail latency SLOs in each QoS bucket compared to baselines. Notably, \sysname's hybrid prioritization smoothly balances between deadline-prioritizing schemes (EDF) and length-prioritizing ones (SRPF), achieving low median latency without drastically increasing tail latency.
\end{itemize}

\begin{figure*}[t]
\hspace{-3cm}
    \centering
    \begin{minipage}{0.3\textwidth}
        \centering
        \includegraphics[width=\textwidth]{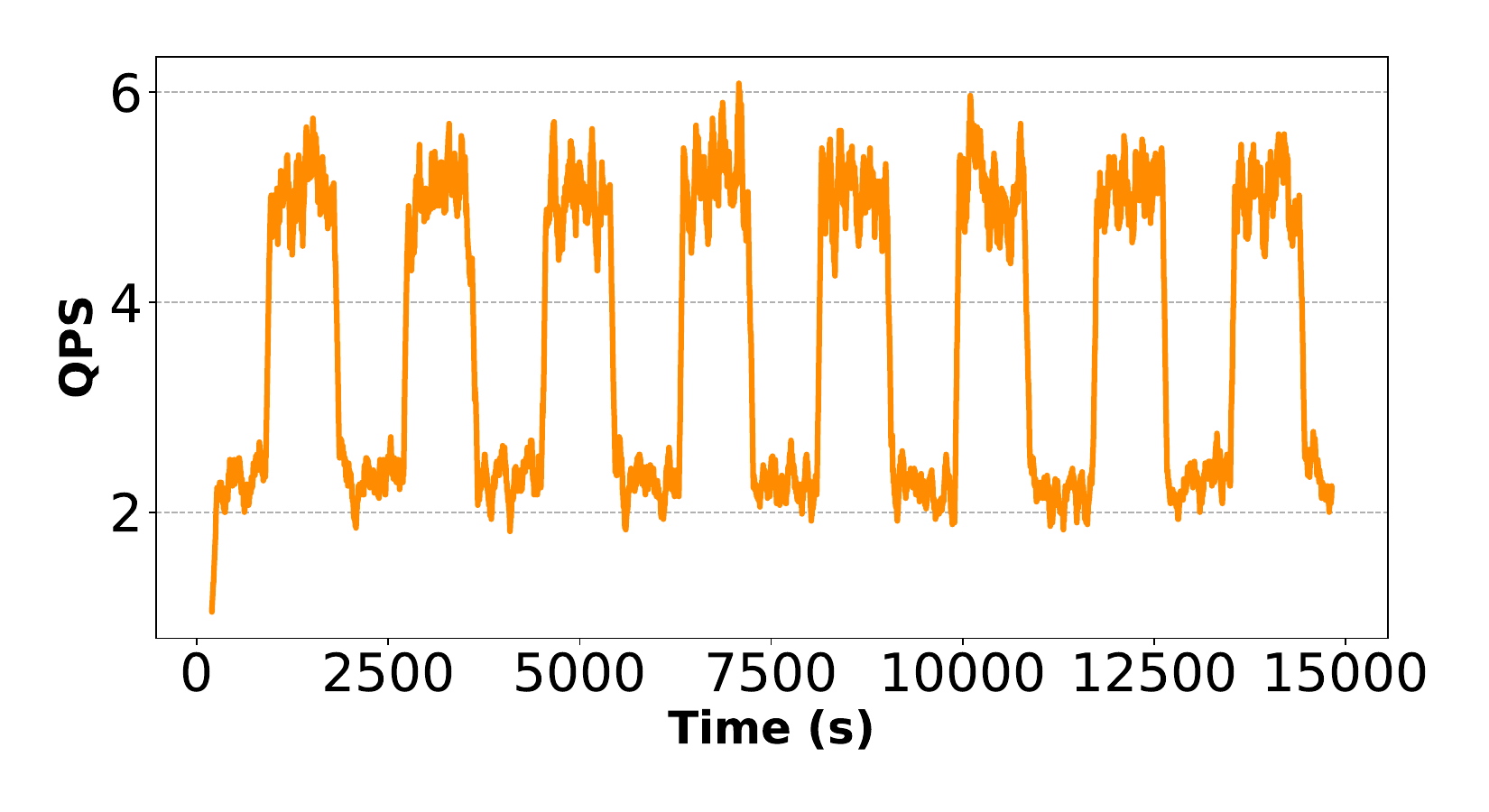}
    \end{minipage}%
    \quad
    \begin{minipage}{0.42\textwidth}
        \centering
        \begin{tabular}{l|c|c|c|c|c}
        \hline
        \multirow{2}{*}{Scheme} & \multicolumn{5}{c}{Violations (\%)} \\ \cline{2-6}
        & Overall & Important & QoS 0 & QoS 1 & QoS 2  \\ \toprule
        Sarathi-FCFS & 81.88 & 81.96 & 97.13 & 89.14 & 59.57 \\
        Sarathi-EDF & 84.12 & 84.09 & 79.3 & 83.27 & 89.77 \\
        \sysname & 8.64 & 0 & 16.03 & 9.98 & 0 \\
        \bottomrule
        \end{tabular}
    \end{minipage}
    \vspace{-1em}
    \caption{Workload with varying QPS and overall deadline violations across different schemes}
    \label{fig:eval:qos_arrival}
\end{figure*}
\begin{figure*}[t]
    \centering
    \includegraphics[width=0.95\linewidth]{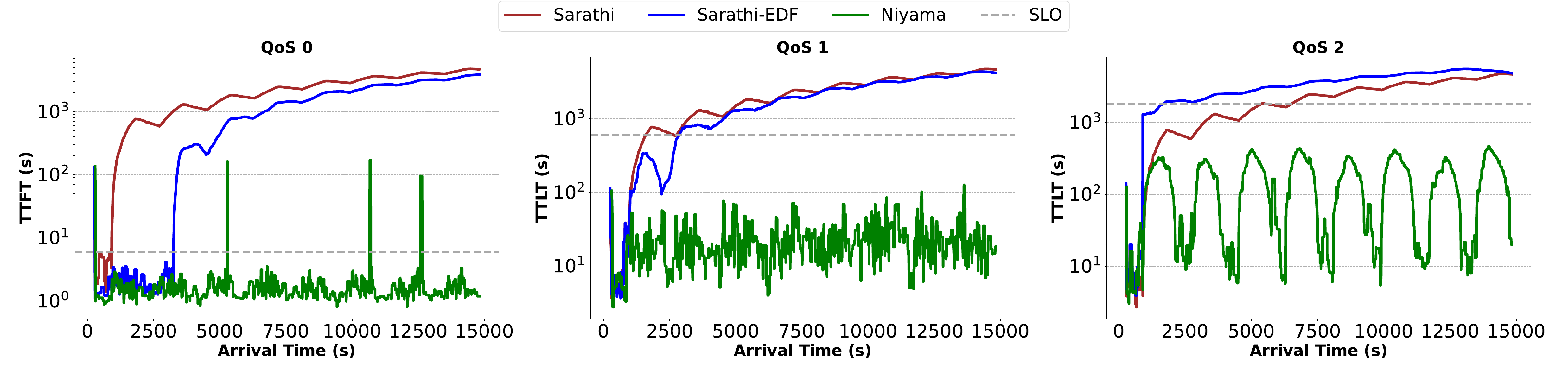}
    \vspace{-1em}
    \caption{Rolling average of p99 latency of all high-priority requests during a dynamic workload with varying request rates}
    \label{fig:eval:vary_qps}
\end{figure*}

\jheading{Deadline violations} For the same workload, \Cref{fig:eval:violations}(a) shows the overall percentage of SLO violations across all requests as load varies. \sysname maintains zero deadline violations for up to 30\% higher load than the next-best scheme, \edf. Even at extreme overloads, \sysname has the fewest deadline violations compared to all other shared-cluster scheduling policies. These lower deadline violation result in the higher goodput we saw in ~\sref{sec:eval:goodput}.
%; virtue of dynamic chunking and hybrid prioritization.

Finally, we analyze whether deadline violations are distributed fairly across request lengths and different QoS buckets. \Cref{fig:eval:violations}(b,c) plot the deadline violations by request length (combined across all QoS buckets). We classify requests as long if their prompt token count is greater than or equal to the 90th percentile, and short otherwise. Our analysis reveals three key patterns:
\begin{itemize}[noitemsep, leftmargin=2em, topsep=0pt]
%\begin{itemize}
    \item \fcfs and \edf violate SLOs for short and long requests at similar rates. These schedulers do not differentiate between request lengths. At high loads when head-of-line blocking occurs, all requests violate SLOs due to a cascade effect.
    \item \srpf shows a very high ratio of violations for long versus short jobs, and ignores all long requests beyond certain load. Even at very low loads ($<$2 QPS), when other schedulers have no deadline violations, \srpf unnecessarily deprioritizes long requests and misses their deadlines. This approach is not only unfair but also counterproductive in real-world settings where request importance doesn't correlate with length.
    \item \sysname achieves balance between these extremes. It does not deprioritize long requests under normal conditions. But during overload, it adjusts the $\alpha$ parameter~\sref{sec:design} to incorporate SRPF-like behavior. This approach allows \sysname to maintain fairness at reasonable loads while gracefully degrading service as load increases.
\end{itemize}

\Cref{fig:eval:violations}d-f plots the split of deadline violations across the three constituent QoS buckets.
We observe that \fcfs first violates requests in the strictest QoS bucket and then continues to the less strict buckets. This happens because \fcfs is deadline unaware, and due to head-of-line blocking, it violates requests with the shortest deadlines when they get blocked by other requests.
\edf equally misses deadlines across all tiers because it treats all requests equally with respect to their individual deadlines.
\srpf shows a pattern similar to \fcfs, violating the strictest tier first due to being deadline unaware. However, it has fewer overall violations by ignoring long jobs, which frees up capacity for the larger proportion of short requests.
On the contrary, \sysname combines the best of these strategies via hybrid prioritization, and achieves fewer overall violations than even \srpf.

\subsection{Latency and SLO violations under Transient Overloads}
\label{sec:eval:overload}
We evaluate whether \sysname can gracefully degrade service during transient overload by running an end-to-end evaluation with diurnal load patterns. Load in the system varies dynamically between low (QPS:2.0) and high (QPS:6) points every 15 minutes over a total of 4 hours as shown in \Cref{fig:eval:qos_arrival}(a). To evaluate \sysname handling of requests with multiple priorities, we mark a random set of 20\% of requests in each QoS bucket as low priority, based on application hints. The remaining 80\% of requests in each bucket are marked as high priority or Important.  

\Cref{fig:eval:qos_arrival}(b) shows the overall deadline violations observed in the system. While the baselines collapse under this load and violate deadlines for all requests, \sysname misses deadlines for no important tasks and only 8.75\% of all requests. This improvement comes from leveraging application hints to perform eager relegation. Additionally, the throughput gains from dynamic chunking and hybrid prioritization help \sysname sustain higher loads.

\Cref{fig:eval:vary_qps} plots the rolling p99 latency (over 60s windows) of all requests in the system for the three QoS buckets. We see that the baseline \fcfs crumbles when the first request burst hits. It cannot recover from the queueing delay and enters request denial mode beyond that point for all classes. While \edf sustains the first burst and absorbs some of it until the second peak, it succumbs to queueing delay beyond this point.
\sysname elegantly handles both high and low load periods, meeting the p99 latency SLOs for a large majority of requests (all important requests and 92\% of all requests). This demonstrates graceful service degradation—proactively dropping a few requests during overload to maintain service levels for the majority, thereby eliminating cascade effects. %In fact, the p50 rolling average for \sysname remains much more uniform and resilient to load. 

\subsection{Ablation Studies}
\label{sec:eval:ablation}
We now examine how each component of our system design affects throughput and SLO violations. For this analysis, we tag all requests as important and evaluate three design elements—dynamic chunking, hybrid prioritization, and eager relegation—starting with the \edf baseline.
\Cref{tbl:eval:ablation} shows that dynamic chunking provides a 20\% boost in throughput, while eager relegation adds another 9\%. The impact of hybrid prioritization appears marginal in the optimal load scenario but becomes significant at high load.
%in varying QPS scenarios as described in \sref{sec:eval:overload}.

\begin{figure}[t]
    \centering
    \includegraphics[width=0.95\linewidth]{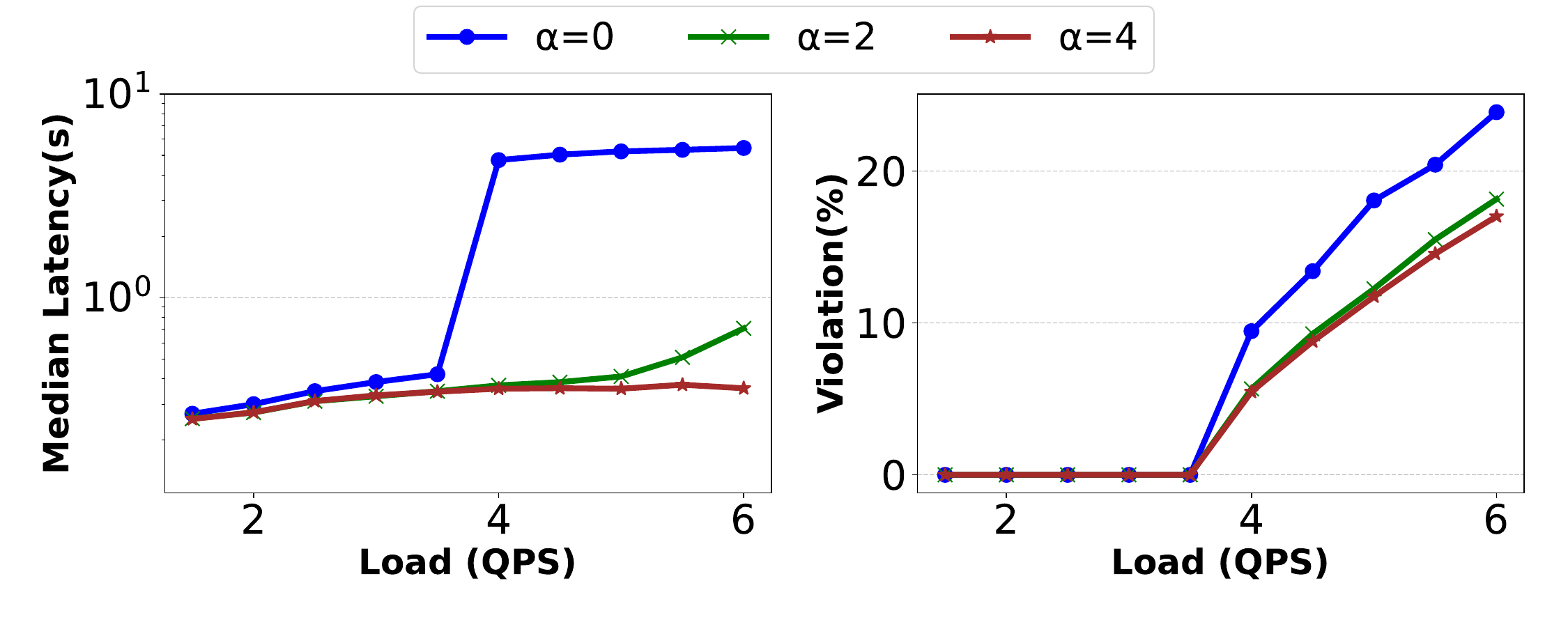}
    \vspace{-1em}
    \caption{Varying the hybrid prioritization parameter}

    \label{fig:eval:c}
\end{figure}

\begin{table}[t!]
    \centering
    \scalebox{0.85}{
    \begin{tabular}{l|cc|cc}
    \multirow{2}{*}{Config} & \multicolumn{2}{c}{Optimal Load} & \multicolumn{2}{|c}{High load (QPS=6)}\\
      & QPS & \% gain & \% viol &  \% impr. \\  \toprule
     %Sarathi-FCFS & 1.5 & \\
     Sarathi-EDF & 2.75 & - & 100 & -\\
     \sysname (DC) & 3.3 & \textcolor{darkgreen}{\textbf{20\%}} & 74 & \textcolor{darkgreen}{\textbf{26\%}}\\
     \sysname (DC+ER) & 3.6 & \textcolor{darkgreen}{\textbf{9\%}} & 26 & \textcolor{darkgreen}{\textbf{68\%}}\\
     \sysname (DC+ER+HP) & 3.65 & \textcolor{darkgreen}{\textbf{1.4\%}} & 16 & \textcolor{darkgreen}{\textbf{32\%}}\\
    \bottomrule
    \end{tabular}}
    \caption{Impact of \sysname's optimizations. (DC:Dynamic Chunking, ER:Eager Relegation, HP:Hybrid Prioritization)}
    \label{tbl:eval:ablation}
    \vspace{-1em}
\end{table}
To further illustrate the impact of hybrid prioritization, \Cref{fig:eval:c} plots the median latency and percentage of deadline violations as we vary system load across three different values of $\alpha$, our hybrid prioritization parameter. As $\alpha$ increases, the system increasingly deprioritizes longer requests. This significantly reduces median latency for all requests but comes at the cost of violating deadlines for most long requests. This demonstrates the importance of tuning this parameter as load increases to strike a balance between low 
 median latency and fair service for long requests.

\section{Related work}
We now place our work in the context of relevant literature.

\jheading{LLM Inference Optimization}
Recent systems research has focused extensively on optimizing LLM inference to reduce latency and increase throughput. Orca~\cite{orca} developed iteration level batching improving the throughput of LLM Inference. vLLM~\cite{vllmsosp} introduced the concept of PagedAttention to manage the KV cache effectively, reducing memory fragmentation and enabling efficient memory sharing across multiple requests. Similarly, Sarathi~\cite{sarathi2024} proposed chunked-prefills and stall-free batching to address the throughput-latency tradeoff. While these advancements have significantly enhanced inference efficiency, they primarily target throughput maximization without explicitly accommodating diverse QoS requirements. Our work builds upon these foundational optimizations to further maximize QoS attainment across multiple service classes.

\begin{comment}
\jheading{LLM Inference Serving}
DistServe ~\cite{distserve2024} Splitwise ~\cite{patel2023splitwise},
use phase-splitting however use FCFS request scheduling. VTC~\cite{sheng2023fairness} develops a scheduler based on client-level fairness. Learning-To-Rank ~\cite{fu2024efficientllmschedulinglearning} ranks requests according to the their output length and prioritizes requests with fewer output tokens left to be generated. However this requires to train a ranking model for which we require to multiplex GPU Resources between the LLM and the ranking model. \jm{I guess this isn't quite relevant to our discussion no?}
\end{comment}

\jheading{Multi-Tenant Serving Systems}
Multi-tenant serving systems for machine learning models have received significant attention in recent years. Clockwork~\cite{clockwork} introduced SLO-aware scheduling for DNN inference assuming a single SLO across all requests, while INFaaS~\cite{infaas} provides an interface for model selection based on accuracy-latency tradeoffs by switching between models. However, these systems don't address the challenge of co-scheduling requests across  different QoS classes on a shared infrastructure and the unique challenges that LLM inference brings therewith.

\jheading{QoS-Aware Scheduling}
Traditional QoS-aware scheduling has been extensively studied in real-time systems literature. Approaches like Earliest Deadline First (EDF)~\cite{edf} and Rate Monotonic Scheduling~\cite{rate_monotonic} provide theoretical foundations for task scheduling with deadlines. In cloud computing, prior work~\cite{paragon_asplos, quasar_asplos} addresses QoS-aware resource management in datacenters by mapping workloads to appropriate resources while minimizing interference. Inspired by the importance of QoS, our work examines the challenges and opportunities in a new domain—LLM serving, with its unique two-phase execution model, high preemption costs, and specialized requirements for handling overload conditions.

\jheading{Graceful service degradation}  
Online systems manage overload through complementary techniques including load balancing~\cite{LoadBalancerBouncer, google-load-balance}, per-client quotas, client side throttling~\cite{google-client-throttle}, resource scheduling~\cite{598025, 10.1145/1125274.1125276}, and graceful degradation~\cite{graceful_blinkdb, google-graceful}. These approaches have been extensively studied in web services, databases, and distributed systems. For example, Bouncer~\cite{Bouncer} introduces an admission control framework for low-latency data systems that makes intelligent decisions about query admission based on response time objectives. To the best of our knowledge, our work is the first to explore these challenges specifically in the context of LLM inference by leveraging its unique characteristics and identifying novel opportunities for maintaining QoS guarantees during overload conditions.
\section{Conclusion}
\label{sec:eval}
We address the challenge of co-scheduling multiple QoS classes in LLM inference serving, and graceful service degradation during overload. We achieve this using three key techniques: (1) dynamic chunking to opportunistically maximize throughput while meeting latency targets, (2) hybrid prioritization to strike a balance between maintaining low median latency and fairness in serving longer requests, and (3) eager relegation to enable graceful service degradataion.
Our evaluation across diverse workloads shows that \sysname significantly improves QoS attainment compared to state-of-the-art LLM serving systems, particularly under high load. As LLMs power more applications with varying performance needs, we believe that techniques supporting multiple QoS classes will become essential for production deployments.

\bibliographystyle{ACM-Reference-Format}
\bibliography{all}

\end{document}